\documentclass[graybox]{svmult}

\usepackage{makeidx}         %
\usepackage{graphicx}        %
\usepackage{multicol}        %
\usepackage[bottom]{footmisc}%
\usepackage{bm}

\usepackage{newtxmath}       %

\usepackage{bbm}%

\usepackage{geometry}
\usepackage{grffile}%
\usepackage{enumitem}%
\usepackage{csquotes}%
\usepackage[
backend=biber,%
natbib=true, %
style=authoryear,%
dashed=false,%
uniquename=init,%
firstinits=true,%
hyperref=true,%
useprefix=true,%
maxcitenames=2,%
maxbibnames=6,%
date=iso8601,%
urldate=iso8601%
]{biblatex}%

\DeclareNameAlias{sortname}{last-first}%
\DefineBibliographyExtras{american}{\DeclareQuotePunctuation{}}%
\renewbibmacro*{volume+number+eid}{%
  \setunit*{\addcomma\space}%
  \printfield{volume}%
  \printfield{number}}
\DeclareFieldFormat*{number}{(#1)}
\DeclareFieldFormat*{title}{#1}%
\DeclareFieldFormat{doi}{%
  \ifhyperref
  {\href{http://dx.doi.org/#1}{\nolinkurl{doi:#1}}}%
  {\nolinkurl{doi:#1}}}%
\renewbibmacro*{in:}{}%
\DeclareFieldFormat{isbn}{ISBN #1}%
\DeclareFieldFormat{pages}{#1}%
\DeclareFieldFormat{url}{\url{#1}}%
\DeclareFieldFormat{urldate}{\mkbibparens{#1}}%
\renewcommand*{\cite}[2][]{\textcite[#1]{#2}}%

\setlist{%
  align=left,%
  labelsep=*,%
  leftmargin=*,%
  topsep=1mm,%
  itemsep=0mm%
}
\newcommand*{\mysquare}{\rule[0.18em]{0.36em}{0.36em}}
\newcommand*{\mytriangle}{\raisebox{0.12em}{\resizebox{0.48em}{0.48em}{$\blacktriangleright$}}}
\newcommand*{\mybar}{\rule[0.32em]{0.62em}{0.08em}}
\newcommand*{\mydot}{\raisebox{0.14em}{\resizebox{0.44em}{!}{$\bullet$}}}
\setlist[itemize,1]{label={\mysquare\ }}%
\setlist[itemize,2]{label={\mytriangle\ }}%
\setlist[itemize,3]{label={\mybar\ }}%
\setlist[itemize,4]{label={\mydot\ }}%
\setlist[enumerate,1]{label=\arabic*.}%
\setlist[enumerate,2]{label=\arabic{enumi}.\arabic*.}%
\setlist[enumerate,3]{label=\arabic{enumi}.\arabic{enumii}.\arabic*.}%

\newcommand{\ou}[3]{%
  \mathrel{%
    \vcenter{\offinterlineskip
      \ialign{##\cr$#1$\cr\noalign{\kern-#3}$#2$\cr}%
    }%
  }%
}
\newcommand*{\omu}[3]{\underset{#3}{\overset{#1}{#2}}}
\newcommand*{\T}{^{\top}}
\newcommand*{\isim}{\omu{\text{\tiny{ind.}}}{\sim}{}}

\newcommand*{\IR}{\mathbbm{R}}

\newcommand*{\U}{\operatorname{U}}

\newcommand*{\N}{\operatorname{N}}

\newcommand*{\ARMA}{\operatorname{ARMA}}
\newcommand*{\GARCH}{\operatorname{GARCH}}

\newcommand*{\ID}{\mathbbm{1}}
\newcommand*{\rd}{\mathrm{d}}

\renewcommand*{\P}{\mathbbm{P}}
\newcommand*{\IE}{\mathbbm{E}} %

\newcommand*{\R}{\textsf{R}}

\newcommand*{\ntrn}{n_{\text{trn}}}
\newcommand*{\nall}{n_{\text{all}}}
\newcommand*{\npth}{n_{\text{pth}}}
\newcommand*{\ngen}{n_{\text{gen}}}
\newcommand*{\nrep}{n_{\text{rep}}}

\makeindex             %

\begin{document}

\title*{Applications of multivariate quasi-random sampling with neural networks}
\author{Marius Hofert, Avinash Prasad and Mu Zhu}
\institute{Marius Hofert \at Department of Statistics and Actuarial Science, University of
	Waterloo, 200 University Avenue West, Waterloo, ON, N2L
	3G1, \email{marius.hofert@uwaterloo.ca}
\and Avinash Prasad \at Department of Statistics and Actuarial Science, University of
Waterloo, 200 University Avenue West, Waterloo, ON, N2L
3G1, \email{a2prasad@uwaterloo.ca}
\and Mu Zhu \at Department of Statistics and Actuarial Science, University of
Waterloo, 200 University Avenue West, Waterloo, ON, N2L
3G1, \email{mu.zhu@uwaterloo.ca}}
\maketitle

\abstract{
  Generative moment matching networks (GMMNs) are suggested for modeling the
  cross-sectional dependence between stochastic processes.  The stochastic
  processes considered are geometric Brownian motions and ARMA--GARCH
  models. Geometric Brownian motions lead to an application of pricing American
  basket call options under dependence and ARMA--GARCH models lead to an application
  of simulating predictive distributions. In both types of applications the
  benefit of using GMMNs in comparison to parametric dependence models is
  highlighted and the fact that GMMNs can produce dependent quasi-random
  samples with no additional effort is exploited to obtain variance reduction.
}

\section{Introduction}
\label{sec:intro}

Given data $\bm{X}_1, \dots, \bm{X}_{\ntrn} \sim F_{\bm{X}}$ in $\IR^d$, a fundamental statistical task is to learn something about $F_{\bm{X}}$ itself. Traditionally, this is done by assuming $F_{\bm{X}}$ to belong to a certain parametric family, say, $F_{\bm{X}}(\cdot; \bm{\theta})$, and estimating the parameter vector $\bm{\theta}$ from data, e.g., by maximum likelihood.

Alternatively, we can learn to generate samples from $F_{\bm{X}}$ directly. Specifically, given a sample $\bm{V}_1, \dots, \bm{V}_{\ngen} \sim F_{\bm{V}}$ in $\IR^d$, where $F_{\bm{V}}$ is a simple distribution (e.g., the independent standard normal or standard uniform), can we learn a generator $G(\cdot)$ such that $G(\bm{V}_1), \dots, G(\bm{V}_{\ngen}) \sim F_{\bm{X}}$? An answer to this question has been provided recently by the machine learning community \parencite{li2015,dziugaite2015}: yes, we can do so using a so-called generative moment matching network (GMMN).

We have found \parencite[see][]{hofertprasadzhu2021} that GMMNs, once trained, also allow us to generate quasi-random samples (QRS) from $F_{\bm{X}}$ ``for free''. This is exciting because we thus have a universal and flexible approach for constructing cross-sectionally dependent QRS from a large variety of different models, including parametric models but especially also empirical ones (for which the true underlying dependence model is unknown). In the present paper, we will first give a quick overview of what GMMNs are, and how they can be used to generate QRS from almost any distribution. We will then focus on two applications: pricing American basket call options, and making probabilistic forecasts for multivariate time series.

Throughout the paper, we rely heavily on the decomposition afforded by Sklar's Theorem \parencite{sklar1959}, namely,
\begin{align*}
  F_{\bm{X}}(\bm{x})=C(F_{X_1}(x_1), \dots, F_{X_d}(x_d)),\quad \bm{x}=(x_1,\dots,x_d)\in\IR^d,
\end{align*}
where $C: [0,1]^d \mapsto [0,1]$ is the unique underlying copula \parencite{embrechtsmcneilstraumann2002,nelsen2006}, and $F_{X_1},\dots,F_{x_d}$ are the continuous marginal distributions of $X_1,\dots,X_d$. This allows us to focus on the problem of generating $\bm{U}_k = G(\bm{V}_k)\sim C$, $k=1,\dots,\ngen$, from which we can simply obtain $\bm{X}_k=F^{-1}_{\bm{X}}(\bm{U}_k) \sim F_{\bm{X}}$, $k=1,\dots,\ngen$, where $F^{-1}_{\bm{X}}(\bm{u})\equiv(F^{-1}_{X_1}(u_1),\dots,F^{-1}_{X_d}(u_d))$.

For both applications we present in this paper, we have not just static data $\bm{X} \sim F_{\bm{X}}$ but realizations of a stretch of a stochastic process $\bm{X}_t$ over time, with
\begin{align}
  \bm{X}_{t_k} = \eta(\bm{Z}_{k}\,|\,\bm{X}_{t_1},\dots,\bm{X}_{t_{k-1}}),\quad k=1,\dots,\ntrn,\label{eq:time-innovation-multi}
\end{align}
where $\bm{Z}_1,\dots,\bm{Z}_{\ntrn}\isim F_{\bm{Z}}$ and $\eta$ is ``decomposable'' into component-wise or marginal functions $\eta_j$, in the sense that, for each $j=1,\dots,d$, we have
\begin{align}
  X_{t_k,j} = \eta_j(Z_{k,j}\,|\,X_{t_1,j},\dots,X_{t_{k-1},j}).\label{eq:time-innovation-uni}
\end{align}
The key structure here is that, for any fixed $j$, $Z_{1,j},\dots,Z_{\ntrn,j}\isim F_{Z_j}$ but, for any fixed $k$, $Z_{k,1},\dots,Z_{k,d}$ are dependent. For each $j$, the function $\eta_j$ allows us to describe the marginal stochastic processes $X_{t,j}$ as a transformation of iid random variables $Z_{t,j}$. In the two applications we consider, $\eta_j$ can be viewed as a function of the conditional mean and variance processes; see Sections~\ref{sec:opt:price:model} and~\ref{sec:mts:model} for details. However, the functional form of $\eta_j$ can be fairly general and the only restriction is that it is invertible.

Conceptually, we may think of a Sklar decomposition at every time point $t$,
\begin{align*}
  F_{\bm{X}_{t_k}}(\bm{x})=C(F_{X_{t_k,1}}(x_{1}), \dots, F_{X_{t_k,d}}(x_{d})),
\end{align*}
where the copula $C$ remains constant over time and hence the same at all $t$,
but the marginal distributions $F_{X_{t_k,1}},\dots,F_{X_{t_k,d}}$ may vary over time. And the operation
\begin{align*}
  F^{-1}_{\bm{X}_{t_k}}(\bm{U}_k) = (F^{-1}_{X_{t_k,1}}(U_{k,1}\,|\,\mathcal{F}_{t_{k-1},1}),\dots,F^{-1}_{X_{t_k,d}}(U_{k,d}\,|\,\mathcal{F}_{t_{k-1},d}))
\end{align*}
is now conditional on the entire history of the process up to and including time $t_{k-1}$, denoted here by the natural filtration $\mathcal{F}_{t_{k-1}}$, where, for any given $s$, $\mathcal{F}_{s,j} =\sigma(\{X_{s',j}: s' \leq s\})$ for all $j=1,\dots,d$. Under \eqref{eq:time-innovation-multi}--\eqref{eq:time-innovation-uni}, the component-wise conditional operation $F^{-1}_{X_{t_k,j}}(\cdot\,|\,\mathcal{F}_{t_{k-1},j})$ is simply
\begin{align*}
  F^{-1}_{X_{t_k,j}}(U_{k,j}\,|\,\mathcal{F}_{t_{k-1},j}) = \eta_j(F^{-1}_{Z_j}(U_{k,j})\,|\,X_{t_1,j},\dots,X_{t_{k-1},j}).
\end{align*}
For American basket call options, each $F_{X_{t_k,j}}(\cdot\,|\,\mathcal{F}_{t_{k-1},j})$ is dictated by an underlying geometric Brownian motion (GBM). For multivariate time series, each $F_{X_{t_k,j}}(\cdot\,|\,\mathcal{F}_{t_{k-1},j})$ is dictated by an underlying $\ARMA$--$\GARCH$ process \parencite{weiss1984,bollerslev1986}.

To price American basket call options as well as to make probabilistic forecasts for multivariate time series, the key lies in repeatedly simulating the time path forward for each $X_{t,j}$. For any given path $i$ and time point $t_k$, this is done by generating
\begin{align}
\hat{X}^{(i)}_{t_k,j}
= \hat{F}^{-1}_{X_{t_k,j}}\big(U^{(i)}_{k,j}\,|\,\mathcal{F}^{(i)}_{t_{k-1},j}\big) = \hat{\eta}_j\big(\hat{F}^{-1}_{Z_j}(U^{(i)}_{k,j})\,|\, \hat{X}^{(i)}_{t_1,j},\dots,\hat{X}^{(i)}_{t_{k-1},j}\big), \label{eq:simulate}
\end{align}%
where $\hat{X}^{(i)}_{s,j} = X_{s,j}$ across all $i$ if time point $s$ is part of the training set.
Typically, each $\bm{U}^{(i)}_{k} = (U^{(i)}_{k,1},\dots,U^{(i)}_{k,d})\T$ is generated from a parametric copula model, whereas we propose to generate it nonparametrically from a GMMN fitted to the training data.

\section{GMMNs, pseudo-random and quasi-random sampling}\label{sec:overview}
As established in Section~\ref{sec:intro}, we focus on generating $\bm{U}_1,\dots,\bm{U}_{\ngen}$ from the underlying copula $C$. To do so, we rely on training data $\hat{\bm{U}}_{1},\dots,\hat{\bm{U}}_{\ntrn}$. These can be obtained by first estimating and then removing the marginal distributions from $\bm{X}_{t_k}$. For our two applications, this is achieved by
\begin{align}
  \hat{U}_{k,j} = \hat{F}_{X_{t_k,j}}(X_{t_k,j}\,|\,\mathcal{F}_{t_{k-1},j}) = \hat{F}_{Z_j}\big(\hat{\eta}^{-1}_j(X_{t_k,j}\,|\,X_{t_1,j},\dots,X_{t_{k-1},j})\big) \label{eq:remove}
\end{align}
under \eqref{eq:time-innovation-multi}--\eqref{eq:time-innovation-uni}.

\subsection{GMMNs}\label{sec:GMMN}
Let $\mathcal{G}$ denote a set of neural networks with a pre-determined architecture.
A GMMN is the solution to the following minimization problem:
\begin{multline}
\underset{G\in\mathcal{G}}{\min}\quad
\frac{1}{\ntrn^2}
\sum_{k=1}^{\ntrn}\sum_{k'=1}^{\ntrn} K(\hat{\bm{U}}_{k},\hat{\bm{U}}_{k'})
- \frac{2}{\ntrn\ngen}
  \sum_{k=1}^{\ntrn}\sum_{k'=1}^{\ngen} K(\hat{\bm{U}}_{k},G(\bm{V}_{k'})) \\
+ \frac{1}{\ngen^2}
  \sum_{k=1}^{\ngen}\sum_{k'=1}^{\ngen} K(G(\bm{V}_{k}),G(\bm{V}_{k'})),
  \label{eq:GMMN-objective}
\end{multline}
where $K(\bm{u},\bm{v})$ is a kernel function, such as the Gaussian or radial basis kernel.
The minimizer of \eqref{eq:GMMN-objective} ensures the distribution of the generated sample $\{G(\bm{V}_{k'})\}_{k'=1}^{\ngen}$ is as close as possible to that of the training sample $\{\hat{\bm{U}}_{k}\}_{k=1}^{\ntrn}$. This is because the criterion being minimized in \eqref{eq:GMMN-objective} is equal to
\begin{align*}
\bigg\Vert
 \frac{1}{\ntrn}\sum_{k=1}^{\ntrn} \varphi(\hat{\bm{U}}_{k})
-\frac{1}{\ngen}\sum_{k'=1}^{\ngen} \varphi(G(\bm{V}_{k'}))
\bigg\Vert^2,
\end{align*}
where $\varphi(\cdot)$ is the implied feature map of $K$ such that $K(\bm{u},\bm{v})=\varphi(\bm{u})\T\varphi(\bm{v})$ and, for the Gaussian kernel and its implied feature map, the two statistics --- $(1/\ntrn)\sum_{k=1}^{\ntrn} \varphi(\hat{\bm{U}}_{k})$ and $(1/\ngen)\sum_{k'=1}^{\ngen} \varphi(G(\bm{V}_{k'}))$ --- contain all empirical moments of the training sample $\{\hat{\bm{U}}_{k}\}_{k=1}^{\ntrn}$ and the generated sample $\{G(\bm{V}_{k'})\}_{k'=1}^{\ngen}$, respectively. This is also where the name GMMN comes from.

Here, we will not go into any more details of how the optimization problem \eqref{eq:GMMN-objective} is actually solved; instead, we simply refer the reader to \cite{li2015}, \cite{dziugaite2015} and \cite{hofertprasadzhu2021}. A very short summary is ``by stochastic gradient descent'' but there are many practical details such as the need for mini-batch optimization, and the use of a mixture (rather than a single) kernel function, and so on.

\subsection{Pseudo-random sampling}
After having trained a GMMN, generating $\npth$ paths of $\ngen$ $d$-dimensional
pseudo-random samples from it can be done as follows.
\begin{algorithm}[Pseudo-random sampling of GMMN-dependent paths]\label{algorithm:GMMN:prng2}
  \begin{enumerate}
  \item Fix $\npth$, the number of paths, and $\ngen$, the number of
    $d$-dimensional samples to be generated for each path.
  \item For $i=1,\dots,\npth$, $k=1,\dots,\ngen$, draw $\bm{V}^{(i)}_k=(V^{(i)}_{k,1},\dots,V^{(i)}_{k,d})\isim F_{\bm{V}}$,
    for example, via $\bm{V}^{(i)}_k=F_{\bm{V}}^{-1}(\bm{U}'^{(i)}_k)$, where $\bm{U}'^{(i)}_k=(U'^{(i)}_{k,1},\dots,U'^{(i)}_{k,d})\isim\U(0,1)^d$.
  \item Return the pseudo-observations of $\bm{U}^{(i)}_k=G(\bm{V}^{(i)}_k)$, $i=1,\dots,\npth$, $k=1,\dots,\ngen$.
  \end{enumerate}
\end{algorithm}

\subsection{Quasi-random sampling}\label{sec:QRS}
As mentioned in Section~\ref{sec:intro}, we have found in
\cite{hofertprasadzhu2021} that GMMNs can preserve low discrepancy --- and thus
achieve a variance reduction effect --- if fed with quasi-random samples. In
Algorithm~\ref{algorithm:GMMN:prng2} above, we can simply replace
$\bm{U}'^{(i)}_k\isim\U(0,1)^d$, for $i=1,\dots,\npth$ and $k=1,\dots,\ngen$, by a randomized quasi-Monte
Carlo (RQMC) point set, such as a randomized Sobol' sequence; see
\cite{hofertprasadzhu2021} for empirical evidence under a great variety of
multivariate distributions (specifically, copulas).
As in \cite[Section~7.3]{lemieux2009}, we generate the RQMC point set in a specific way. To this end, let
$d^\ast=\ngen\cdot d$. We then generate an RQMC point set
$\tilde{P}_{\npth}=\{\tilde{\bm{v}}_1,\dots,\tilde{\bm{v}}_{\npth}\}$ of $\npth$
$d^\ast$-dimensional points, resulting in an $(\npth,d^\ast)$-matrix, whose
columns are blocked in $\ngen$ groups of size $d$ each to form $\npth$ paths of
$\ngen$ $d$-dimensional quasi-random samples from the trained GMMN.  We thus
obtain the following algorithm.
\begin{algorithm}[Quasi-random sampling of GMMN-dependent paths]\label{algorithm:GMMN:qrng2}
  \begin{enumerate}
  \item Fix $\npth$, the number of paths, and $\ngen$, the number of $d$-dimensional
    samples to be generated for each path. Furthermore, set $d^\ast=\ngen\cdot d$.
  \item Compute a $d^{*}$-dimensional RQMC point set
    $\tilde{P}_{\npth}=\{\tilde{\bm{v}}_1,\dots,\tilde{\bm{v}}_{\npth}\}$, for
    example, as a randomized Sobol' sequence.
  \item Compute $\bm{V}^{(i)}_{k}=F^{-1}_{\bm{V}}(\tilde{v}_{i,(k-1)d+1},\dots,\tilde{v}_{i,kd})$,
    $i=1,\dots,\npth$, $k=1,\dots,\ngen$.
  \item Return the pseudo-observations of $\bm{U}^{(i)}_k=G(\bm{V}^{(i)}_k)$, $i=1,\dots,\npth$, $k=1,\dots,\ngen$.
  \end{enumerate}
\end{algorithm}

\section{American basket option pricing}\label{sec:opt:price}

In this section, we demonstrate the usefulness of GMMNs and the dependent quasi-random samples they can generate for pricing American basket call options.

\subsection{Model}\label{sec:opt:price:model}
The basket portfolio consists of $d$ assets whose
prices $X_{t,1},\dots,X_{t,d}$ %
follow geometric Brownian motions. %
With risk-neutral drift, $(X_{t,j})_{t\ge 0}$ can be represented as
\begin{align}
  X_{t,j}=X_{0,j}\exp(Y_{t,j})\quad\text{for}\quad Y_{t,j}=(r-\sigma_j^2/2)t+\sigma_jW_{t,j},\label{eq:GBM}
\end{align}
where $r$ is the risk-free interest rate, $\sigma_j$ is the volatility
parameter of the $j$th asset (quantified as the standard deviation of $Y_{t,j}$
over one unit of time) and $W_{t,j}$ is a standard Wiener process. A discretization
of $W_{t,j}$ on $0=t_0<t_1<\dots<t_{\ntrn}$ is given by
\begin{align}
  W_{t_k,j}=\sum_{l=1}^k\sqrt{t_{l}-t_{l-1}}Z_{l,j},\quad Z_{1,j},\dots,Z_{\ntrn,j}\isim\N(0,1),\label{eq:W:discr}
\end{align}
and thus a discretization of \eqref{eq:GBM} is given by %
\begin{align}
  X_{t_k,j}=X_{t_0,j}\exp(Y_{t_k,j})\quad\text{for}\quad Y_{t_k,j}=(r-\sigma_j^2/2)t_k+\sigma_j\sum_{l=1}^k\sqrt{t_{l}-t_{l-1}}Z_{l,j};\label{eq:GBM:discr}
\end{align}
this is in line with Equation~\eqref{eq:time-innovation-uni}.
Dependence between $X_{t_k,1},\dots,X_{t_k,d}$ is introduced by making the increments $Z_{k,1},\dots,Z_{k,d}$ dependent either through
a parametric copula $C_{\text{PM}}$ (in which case we speak of a \emph{copula--GBM
  model} for the joint stock price process
$\bm{X}_{t_k}=(X_{t_k,1},\dots,X_{t_k,d})$, $k=0,\dots,\ntrn$) or through a GMMN (in which case we speak of a \emph{GMMN--GBM model}).

\subsection{Estimation}
Suppose that for each $j=1,\dots,d$, we have $\ntrn+1$ realizations
$X_{t_k,j}$, $k=0,\dots,\ntrn$.  We fix the risk-free interest rate to be $r$ and estimate $\sigma_j$ as the sample standard
deviation $\hat{\sigma}_j$ of the log-returns
$Y_{t_k,j}=\log(X_{t_k,j}/X_{t_{k-1},j})$, $k=1,\dots,\ntrn$. We then recover
the realizations $\hat{Z}_{1,j},\dots,\hat{Z}_{\ntrn,j}$ from $X_{t_0,j},\dots,X_{t_{\ntrn},j}$ via \eqref{eq:W:discr} and \eqref{eq:GBM:discr} by
\begin{align*}
  \hat{W}_{t_k,j}&=\frac{1}{\hat{\sigma}_j}\biggl(\log\biggl(\frac{X_{t_k,j}}{X_{t_0,j}}\biggr) - \biggl(r-\frac{\hat{\sigma}_j^2}{2}\biggr)t_k\biggr),\quad k=0,\dots,\ntrn,\\
  \hat{Z}_{k,j}&=\frac{\hat{W}_{t_k,j}-\hat{W}_{t_{k-1},j}}{\sqrt{t_k-t_{k-1}}},\quad k=1,\dots,\ntrn,
\end{align*}
a process we refer to as \emph{deBrowning} (analogously to deGARCHing known for $\ARMA$--$\GARCH$ processes);
this is in line with Equation~\eqref{eq:remove}.
The cross-sectional dependence of the iid $\hat{\bm{Z}}_k=(\hat{Z}_{k,1},\dots,\hat{Z}_{k,d})$,
$k=1,\dots,\ntrn$, with supposedly standard normal margins is then
modeled based on the \emph{pseudo-observations}
\begin{align}
  \hat{U}_{k,j}=\frac{\hat{R}_{k,j}}{\ntrn+1},\quad k=1,\dots,\ntrn,\ j=1,\dots,d,\label{eq:pobs}
\end{align}
where $\hat{R}_{k,j}$ denotes the rank of $\hat{Z}_{k,j}$ among
$\hat{Z}_{1,j}\dots,\hat{Z}_{\ntrn,j}$. Note that using the pseudo-observations instead of
assuming standard normality of the margins of $\hat{Z}_k=(\hat{Z}_{k,1},\dots,\hat{Z}_{k,d})$
reduces the effect of a potential misspecification of the margins on the
estimation of the cross-sectional dependence; see \cite{genest2010}.  As
cross-sectional dependence model for the distribution of
$\hat{\bm{U}}_k=(\hat{U}_{k,1},\dots,\hat{U}_{k,d})$, $k=1,\dots,\ntrn$, we use
either a fitted parametric copula $\hat{C}_{\text{PM}}$ or a trained GMMN
$G: \IR^d \rightarrow [0,1]^{d}$. Analogously to $\hat{C}_{\text{PM}}$, we
denote the copula of the samples generated from the trained GMMN by
$\hat{C}_{\text{NN}}$ and view them as an approximation to the target dependence
structure of $\hat{\bm{U}}_1,\dots,\hat{\bm{U}}_{\ntrn}$.  In our option pricing
application, we compare the trained GMMN with fitted Clayton, normal and $t$
copulas --- with unstructured correlation matrices for the latter two. We also
include the independence copula as benchmark.

\subsection{Simulation}
For simulating the dependent asset prices, samples from $\hat{C}_{\text{NN}}$
and $\hat{C}_{\text{PM}}$ are mapped to $\N(0,1)$ margins to obtain samples from
the joint increment distribution and thus, after undoing deBrowning, to obtain
samples from the dependent asset prices. The following algorithm describes these steps
for simulating $\npth$-many paths of $\bm{X}_{t_k}=(X_{t_k,1},\dots,X_{t_k,d})$, $k=0,\dots,\ngen$,
for our newly proposed GMMN--GBM model.
\begin{algorithm}[Pseudo- and quasi-random sampling of GMMN-GBM paths]\label{alg:GMMN:GBM:sim}
  \begin{enumerate}
  \item Fix $\npth$, the number of paths, and $\ngen$, the number (or total time steps)
    of $d$-dimensional samples to be generated for each path. Furthermore, fix the risk-free
    interest rate $r$, the estimated asset price volatilities $\hat{\sigma}_j$,
    $j=1,\dots,d$, and the initial asset prices
    $\hat{\bm{X}}_{t_0}^{(i)}=\bm{X}_{t_{\ntrn}}$, $i=1,\dots,\npth$.
  \item Generate $\bm{U}^{(i)}_k=(U^{(i)}_{k,1},\dots,U^{(i)}_{k,d})$, $i=1,\dots,\npth$,
    $k=1,\dots,\ngen$, according to Algorithm~\ref{algorithm:GMMN:prng2} (for
    pseudo-random samples) or Algorithm~\ref{algorithm:GMMN:qrng2} (for
    quasi-random samples).
  \item Compute the increments $Z_{k}^{(i)}=(Z_{k,1}^{(i)},\dots,Z_{k,d}^{(i)})$
    with $Z_{k,j}^{(i)}=\Phi^{-1}(U_{k,j}^{(i)})$ for $i=1,\dots,\npth$,
    $k=1,\dots,\ngen$, where $\Phi^{-1}$ is the $\N(0,1)$ quantile function.
  \item Compute $\hat{\bm{X}}_{t_k}^{(i)}=(\hat{X}_{t_k,1}^{(i)},\dots,\hat{X}_{t_k,d}^{(i)})$ with
    \begin{align*}
      \hat{X}_{t_k,j}^{(i)}=\hat{X}_{t_0,j}^{(i)}\exp\biggl((r-\hat{\sigma}_j^2/2)t_k+\hat{\sigma}_j\sum_{l=1}^k\sqrt{t_{l}-t_{l-1}}Z_{l,j}^{(i)}\biggr)
    \end{align*}
    for $i=1,\dots,\npth$, $k=1,\dots,\ngen$; see \eqref{eq:GBM:discr}.
  \item Return $\hat{\bm{X}}_{t_k}^{(i)}$, $i=1,\dots,\npth$, $k=1,\dots,\ngen$.
  \end{enumerate}
\end{algorithm}
The dependent asset price processes then serve as inputs for computing the
present value of the American basket call option considered.

\subsection{Application}
To price an American basket call option, we assume the option can be exercised at
$t_1,\dots,t_{\ngen}$ and that $t_{\ngen}=T$ is the maturity (or expiry) of the option
contract. We are then interested in estimating
\begin{align}
  \max_{k=1,\dots,\ngen}\IE(\exp(-rt_k)H(\bm{X}_{t_k},t_k)),\label{eq:option}
\end{align}
where
$H(\bm{X}_{t_k},t_k)= \max \bigr\{\bigl((1/d)\sum_{j=1}^{d}X_{t_k,j}\bigr)-K,
0\bigr\}$ is the payoff function for the American basket call option with
strike price $K$; note that the expectation in~\eqref{eq:option} is with
respect to the risk-neutral measure.  To find a solution of \eqref{eq:option},
we follow the dynamic programming principle and traverse each simulated path of
the underlying asset price process backwards in time (starting at maturity) while making hold/exercise decisions at each time point. To make
such decisions along the $i$th path, we need to compare the \emph{exercise value}
$H(\bm{X}_{t_k}^{(i)},t_k)$ with the expected \emph{continuation value}. To this end, we
work with the least squares Monte Carlo algorithm of \cite{longstaff2001}
according to which the expected continuation value is estimated by regressing the
realized option payoffs from continuation on basis functions of the basket price.
As basis functions we use the first three weighted Laguerre polynomials
\begin{align*}
  L_{0}(x)=e^{-x/2}, \quad L_{1}(x)=e^{-x/2}(1-x^2/2),\quad L_{2}(x)=e^{-x/2}(1-2x +x^2/2),
\end{align*}
where $x$ is the current asset price, and we also use their corresponding three cross-products
$L_{0}(x)L_{1}(x)$, $L_{0}(x)L_{2}(x)$ and $L_{1}(x)L_{2}(x)$. Algorithm~\ref{alg:lsm:basket}
provides the details; see also \cite{longstaff2001}.
\begin{algorithm}[Least squares Monte Carlo for American basket call option pricing]\label{alg:lsm:basket}
  \begin{enumerate}
  \item Simulate asset price paths $\hat{\bm{X}}_{t_k}^{(i)}$, $i=1,\dots,\npth$,
    $k=1,\dots,\ngen$, using Algorithm~\ref{alg:GMMN:GBM:sim} based on either
    GMMN pseudo-random samples or GMMN quasi-random samples.
  \item Compute the value of the American basket call option at maturity:
    $\hat{V}^{(i)}_{t_{\ngen}}= H(\hat{\bm{X}}^{(i)}_{t_{\ngen}}, t_{\ngen})$, $i=1,\dots,\npth$.
  \item For $k=\ngen-1,\dots,1$ do:
    \begin{enumerate}
    \item Compute the discounted values
      $\hat{V}^{(i)}_{t_k}=\exp(-r(t_k-t_{k-1}))\hat{V}^{(i)}_{t_{k+1}}$,
      $i=1,\dots,\npth$.
    \item Compute the basket price
      $\bar{\hat{X}}^{(i)}_{t_{k}}=\frac{1}{d}\sum_{j=1}^{d}\hat{X}^{(i)}_{t_{k},j}$,
      $i=1,\dots,\npth$.
    \item Compute the design matrix $D_{t_{k}}\in \IR^{\npth \times 7}$ with
      $i$th row given by $(1, L_{0}(\bar{\hat{X}}^{(i)}_{t_{k}})$,
      $L_{1}(\bar{\hat{X}}^{(i)}_{t_{k}})$, $L_{2}(\bar{\hat{X}}^{(i)}_{t_{k}})$,
      $L_{0}(\bar{\hat{X}}^{(i)}_{t_{k}})L_{1}(\bar{\hat{X}}^{(i)}_{t_{k}})$,
      $L_{0}(\bar{\hat{X}}^{(i)}_{t_{k}})L_{2}(\bar{\hat{X}}^{(i)}_{t_{k}})$,
      $L_{1}(\bar{\hat{X}}^{(i)}_{t_{k}})L_{2}(\bar{\hat{X}}^{(i)}_{t_{k}}))$,
      $i=1,\dots,\npth$.
    \item Regress $\hat{V}^{(i)}_{t_k}$, $i=1,\dots,\npth$, on $D_{t_{k}}$ and use
      the fitted values of the regression as continuation values, i.e.,
      $\hat{C}^{(i)}_{t_k}=D_{t_{k}}(D_{t_{k}}'D_{t_{k}})^{-1}D_{t_{k}}'\hat{V}^{(i)}_{t_k}$,
      $i=1,\dots,\npth$. Adjust the continuation values for bias
      by setting $\hat{C}^{(i)}_{t_k}= \max\{\hat{C}^{(i)}_{t_k},0\}$, $i=1,\dots,\npth$; see
      \cite{longstaff2001}.
    \item Compute the exercise values
      $\hat{E}^{(i)}_{t_{k}}=H\big(\hat{\bm{X}}^{(i)}_{t_{k}},t_{k}\big)$,
      $i=1,\dots,\npth$.
    \item Set $\hat{V}^{(i)}_{t_{k}}=\hat{E}^{(i)}_{t_{k}}$ for all $i=1,\dots,\npth$ such
      that $\hat{E}^{(i)}_{t_{k}} \geq \hat{C}^{(i)}_{t_k}$.
    \end{enumerate}
  \item Compute the American basket call option price as
    $e^{-r(t_1-t_0)}\frac{1}{\npth}\sum_{i=1}^{\npth}\hat{V}^{(i)}_{t_1}$.
  \end{enumerate}
\end{algorithm}

As data application we consider daily adjusted closing prices of 10 S\&P~500
constituents from 1995-01-01 to 2015-12-31. The selected constituents include
three stocks from the information technology sector (Intel Corp.\ (INTC), Oracle
Corp.\ (ORCL) and International Business Machines Corp.\ (IBM)), three stocks
from the financial sector (Capital One Financial Corp.\ (COF), JPMorgan Chase \&
Co.\ (JPM) and American International Group Inc (AIG)) and four stocks from the
industrial sector (3M Company (MMM), Boeing Company (BA), General Electric (GE)
and Caterpillar Inc.\ (CAT)). In addition, we also consider sub-portfolios of
these constituents with dimensions $d=5$ (consisting of INTC, ORCL, IBM, COF and
AIG) and $d=3$ (consisting of INTC, IBM and AIG). The data used for this
application can be obtained from the \R\ package \texttt{qrmdata}. We choose the
risk-free interest rate to be $r=0.0005$ (annualized), in line with the US
treasury bond yield rates at the onset of the option contract for maturities
similar to the lengths of the contracts considered; the maturities in our
pricing exercise range from one day to 100
days.  %
When selecting the strike price $K$ for the various basket options with
different maturities, we try to ensure that the values of these basket call
options are not too close to zero by selecting $K$ to be close to the current
market value of the basket --- specifically, 101\% of the basket value at time
point $t_{\ntrn}$.

To model the pseudo-observations \eqref{eq:pobs} of each of the the three
portfolios considered, we use parametric copulas $C_{\text{PM}}$ that are known
to capture the dependence between financial return series well, such as the
normal and the $t$. We also consider a Clayton copula which is lower tail
dependent. The choice of the normal and the $t$ is also because we are able to
generate \emph{quasi}-random samples from them, while quasi-random sampling is
not readily available for all parametric copula models; see
\cite{hofertprasadzhu2021}. This allows us to compare the variance reduction
factors achieved when pricing American basket call options with the GMMN versus
parametric copulas such as the normal and the $t$.  For GMMNs, we use the same
architecture and choice of hyperparameters as described in
\cite{hofertprasadzhu2021}. %
All parametric copulas are fitted using the maximum pseudo-likelihood method;
see \cite[Section~4.1.2]{hofertkojadinovicmaechleryan2018}.
For a detailed comparison of run time measurements between GMMNs and parametric copulas in the contexts of training (or model fitting), as well as for pseudo-random and quasi-random sampling, refer to  \cite[Appendix~B]{hofertprasadzhu2021}.

\begin{figure}[htbp]
\centering
\includegraphics[width=0.33\textwidth]{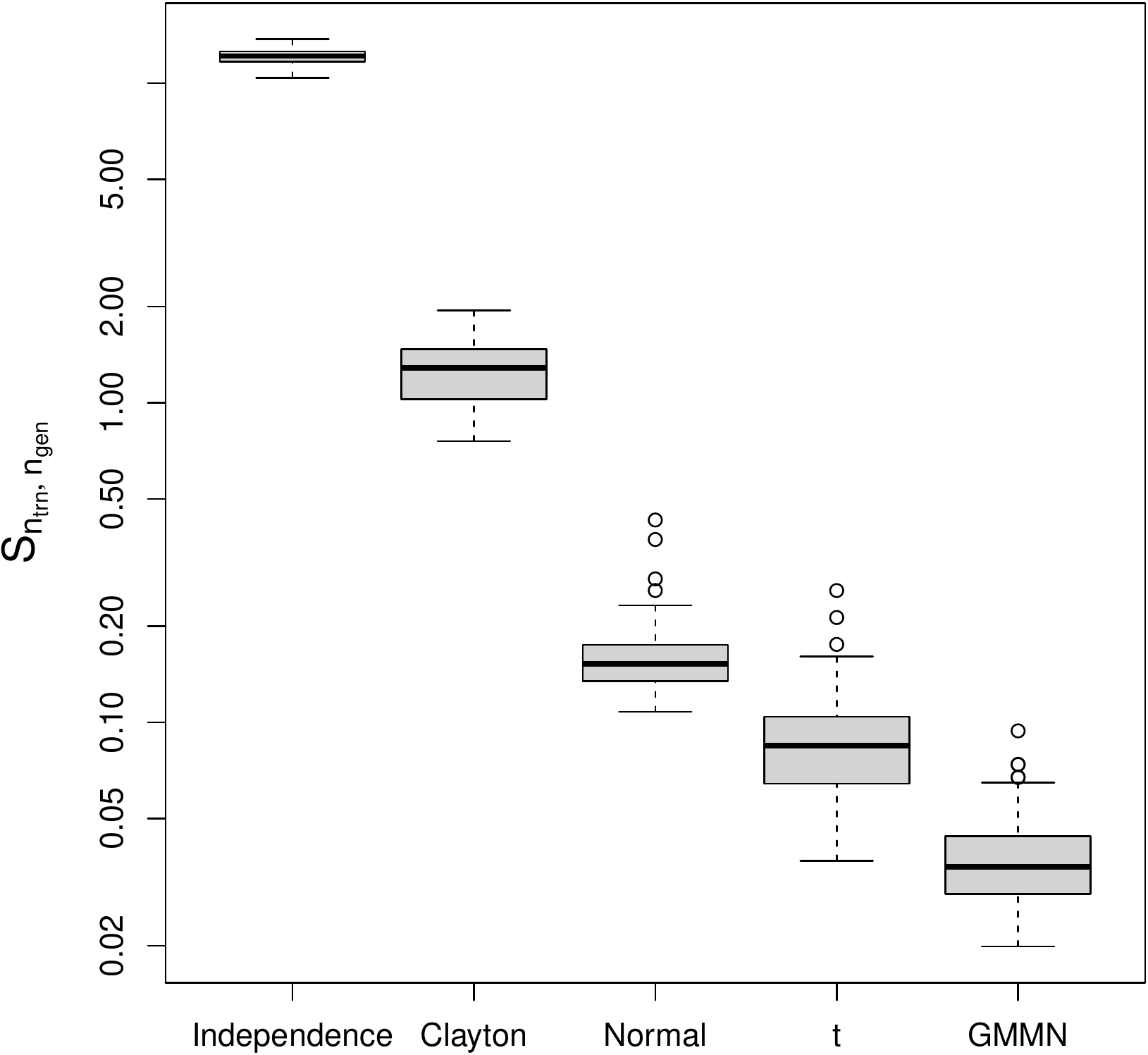}\hfill
\includegraphics[width=0.33\textwidth]{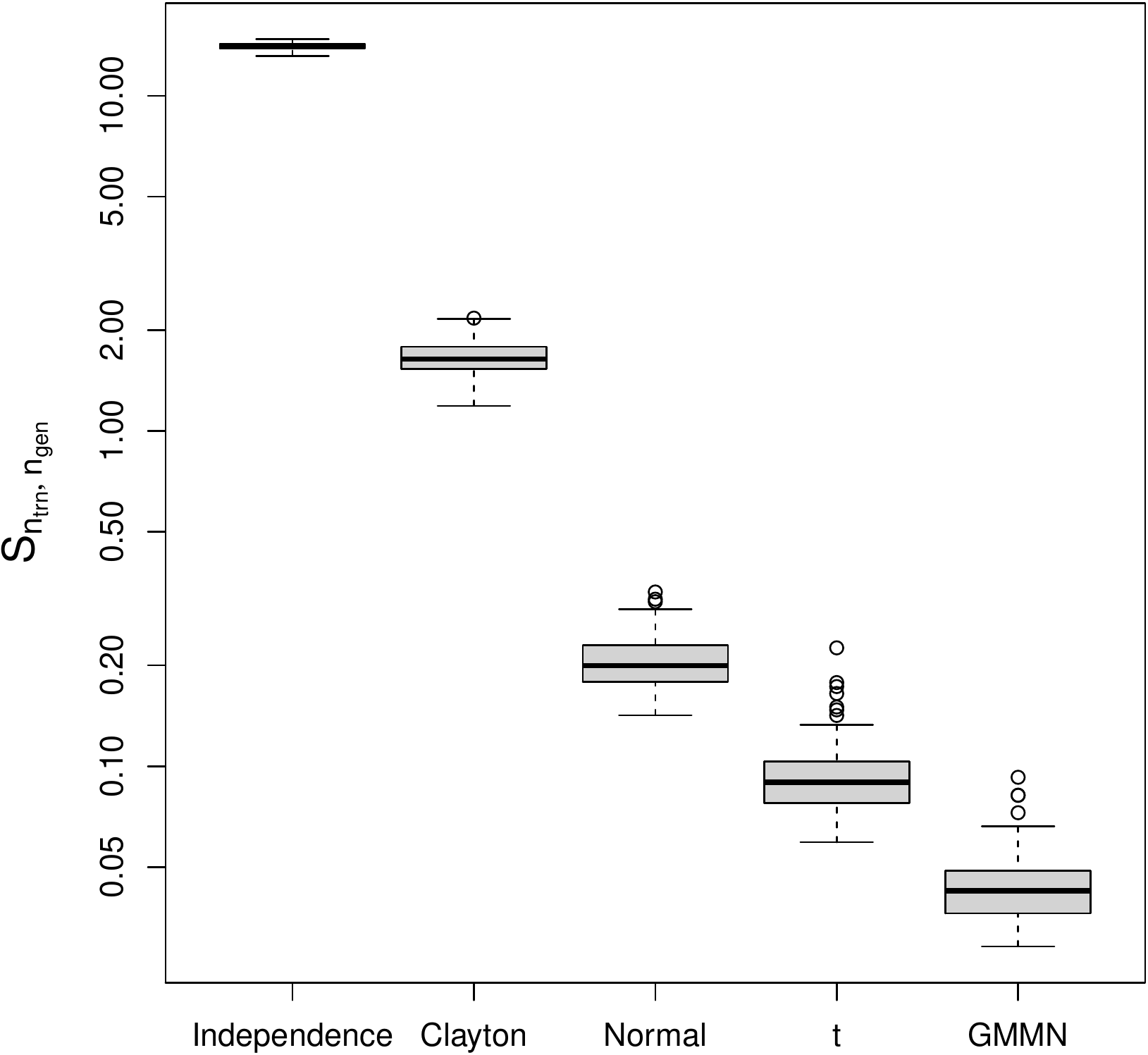}\hfill
\includegraphics[width=0.33\textwidth]{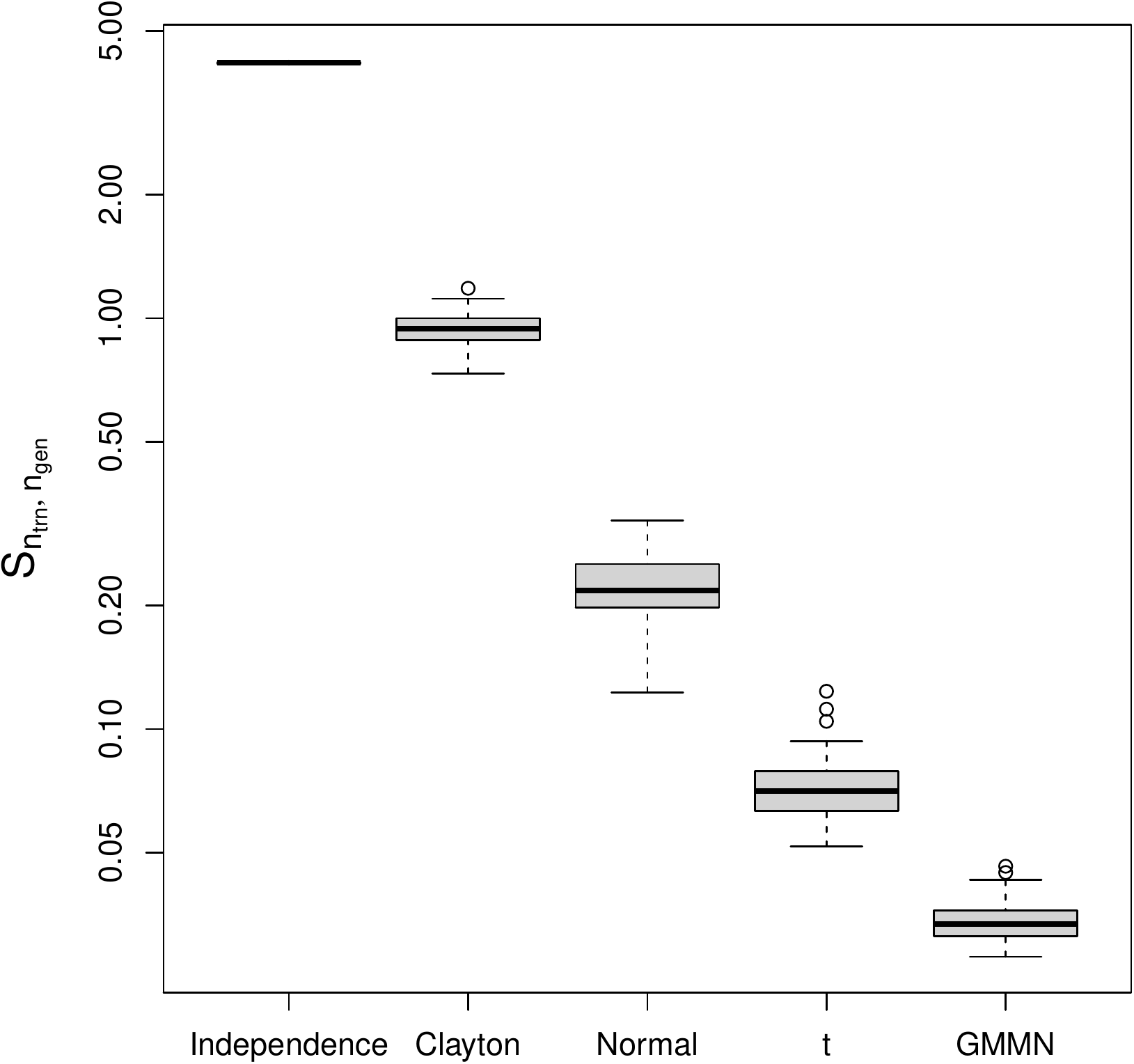}
\caption{Box plots based on $\nrep=100$ realizations of
    $S_{\ntrn,\ngen}$ computed for portfolios of dimensions $d=3$ (left), $d=5$
    (middle) and $d=10$ (right) with training sample size $\ntrn=5287$. For each
    fitted dependence model we generate a pseudo-random sample of size
    $\ngen=10\,000$. From these box plots, we can see that the GMMNs provide a
    much better fit than well-known parametric copulas for all three portfolios
    considered.}\label{fig:gof:option}
\end{figure}

To evaluate the fit of a dependence model, we use a Cram\'{e}r-von-Mises type
test statistic introduced by \cite{remillardscaillet2009} to assess
the equality of two empirical copulas; see \cite{hofertprasadzhu2021}. This statistic is defined as
\begin{align}
  S_{\ntrn,\ngen} = \frac{1}{\sqrt{\frac{1}{\ntrn}+\frac{1}{\ngen}}}\int_{[0,1]^d} \bigl(C_{\ntrn}(\bm{u})-C_{\ngen}(\bm{u})\bigr)^2\,\rd \bm{u}, \label{eq:cvm}
\end{align}
where $C_{\ntrn}(\bm{u})$ is the empirical copula of the $\ntrn$
pseudo-observations used to fit the dependence model (see \eqref{eq:pobs}) and
$C_{\ngen}(\bm{u})$ is the empirical copula of the $\ngen$ samples generated
from the fitted dependence model (either $\hat{C}_{\text{PM}}$ or
$\hat{C}_{\text{NN}}$). Figure~\ref{fig:gof:option} shows box plots of
$S_{\ntrn,\ngen}$ for the different models based on $\nrep=100$ repetitions; see
\cite[Section~2]{remillardscaillet2009} for how to evaluate $S_{\ntrn,\ngen}$.
As we can see, GMMNs provide the best fit according to $S_{\ntrn,\ngen}$ across
all dimensions considered.

\begin{figure}[htbp]
\centering
\includegraphics[width=0.33\textwidth]{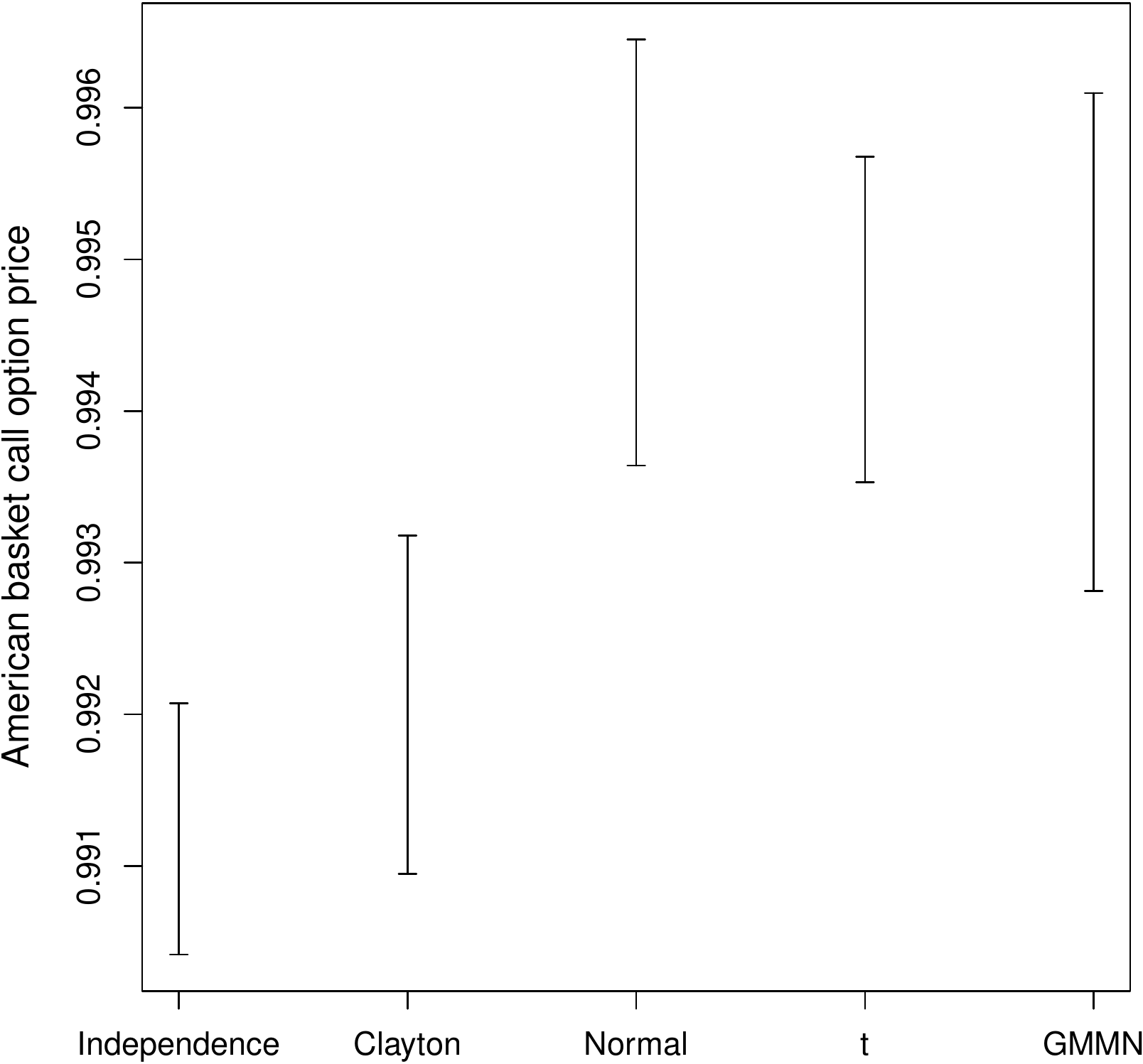}\hfill
\includegraphics[width=0.33\textwidth]{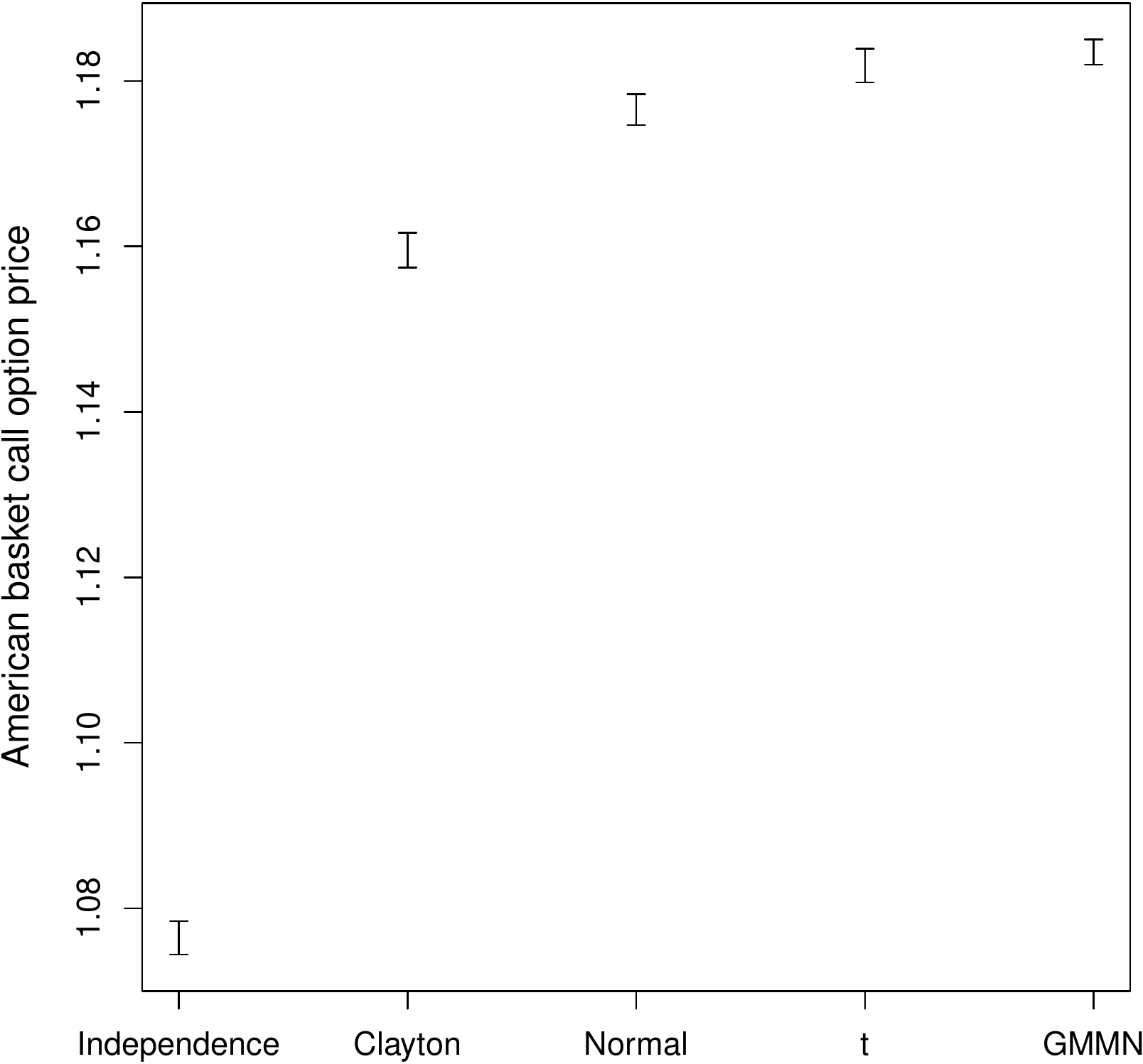}\hfill
\includegraphics[width=0.33\textwidth]{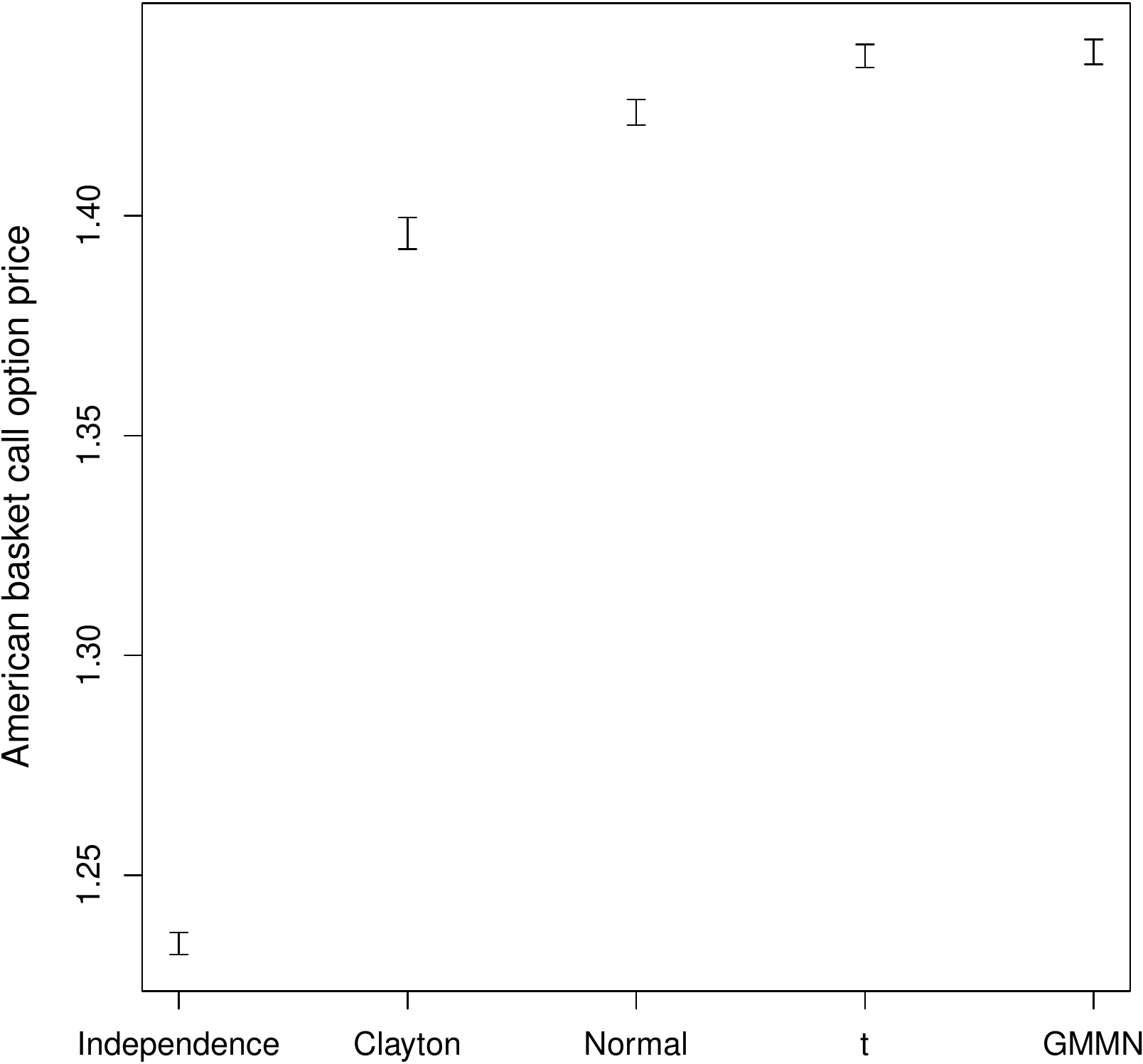}\\[2mm]
\includegraphics[width=0.33\textwidth]{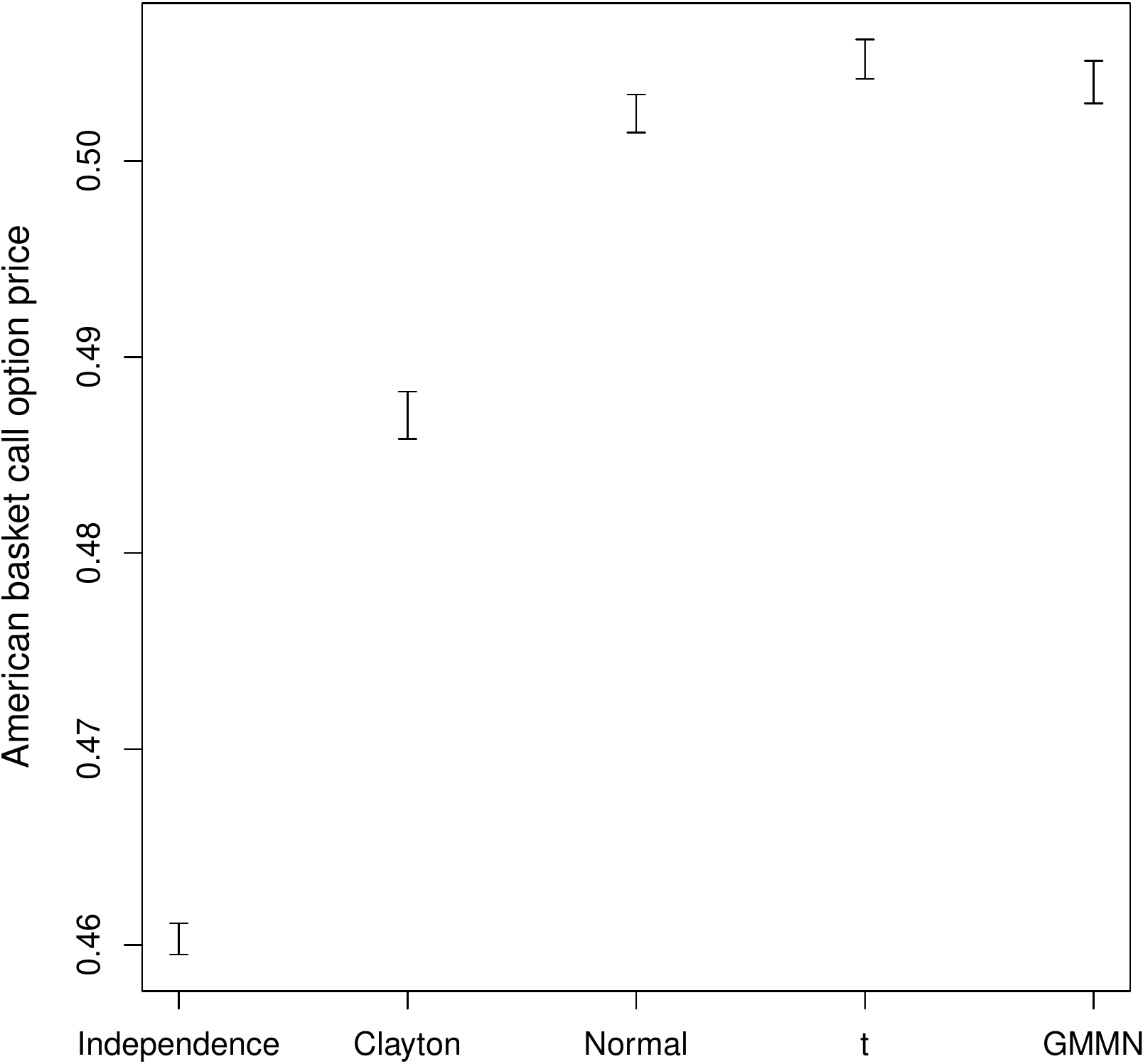}\hfill
\includegraphics[width=0.33\textwidth]{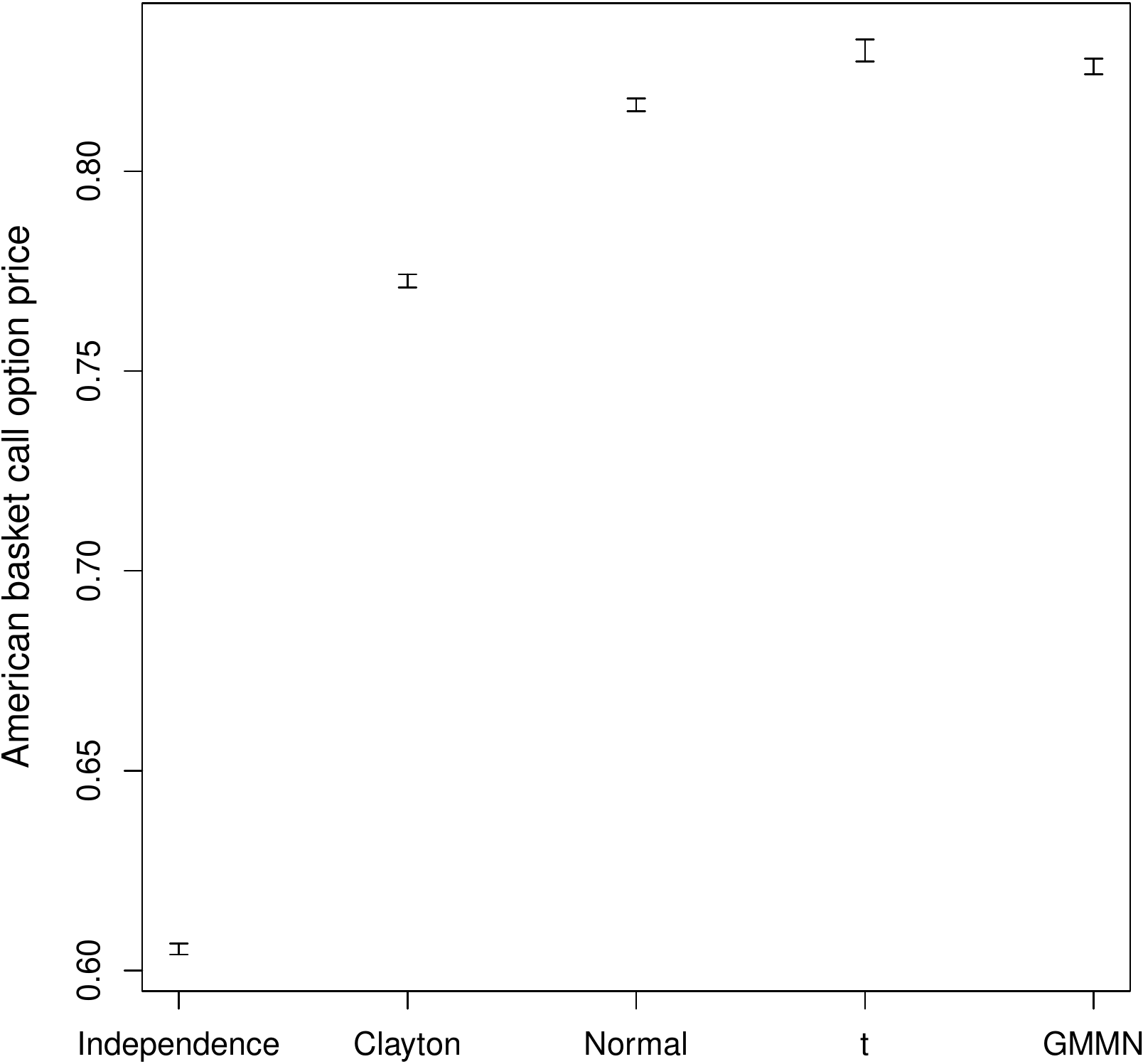}\hfill
\includegraphics[width=0.33\textwidth]{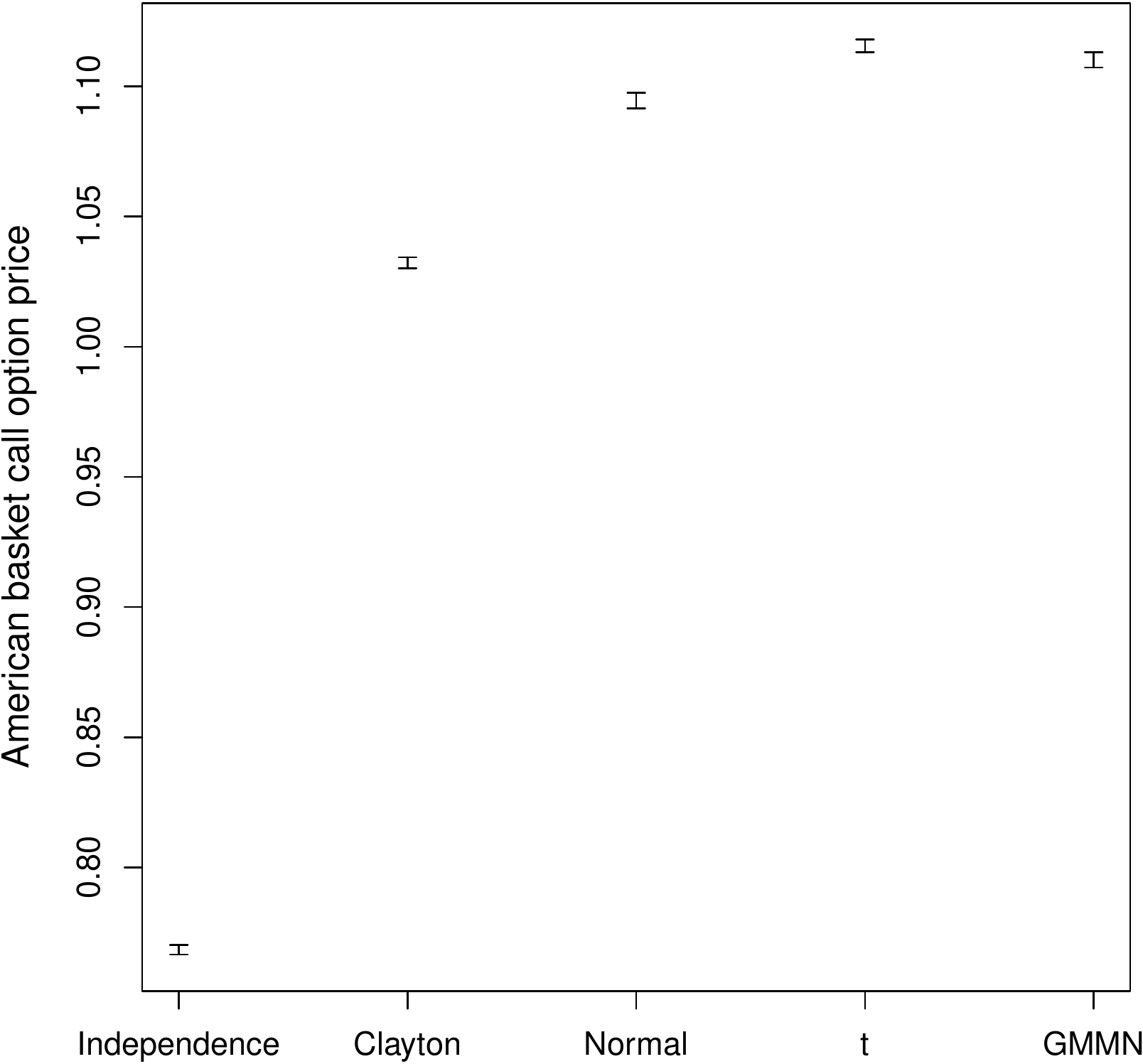}\\[2mm]
\includegraphics[width=0.33\textwidth]{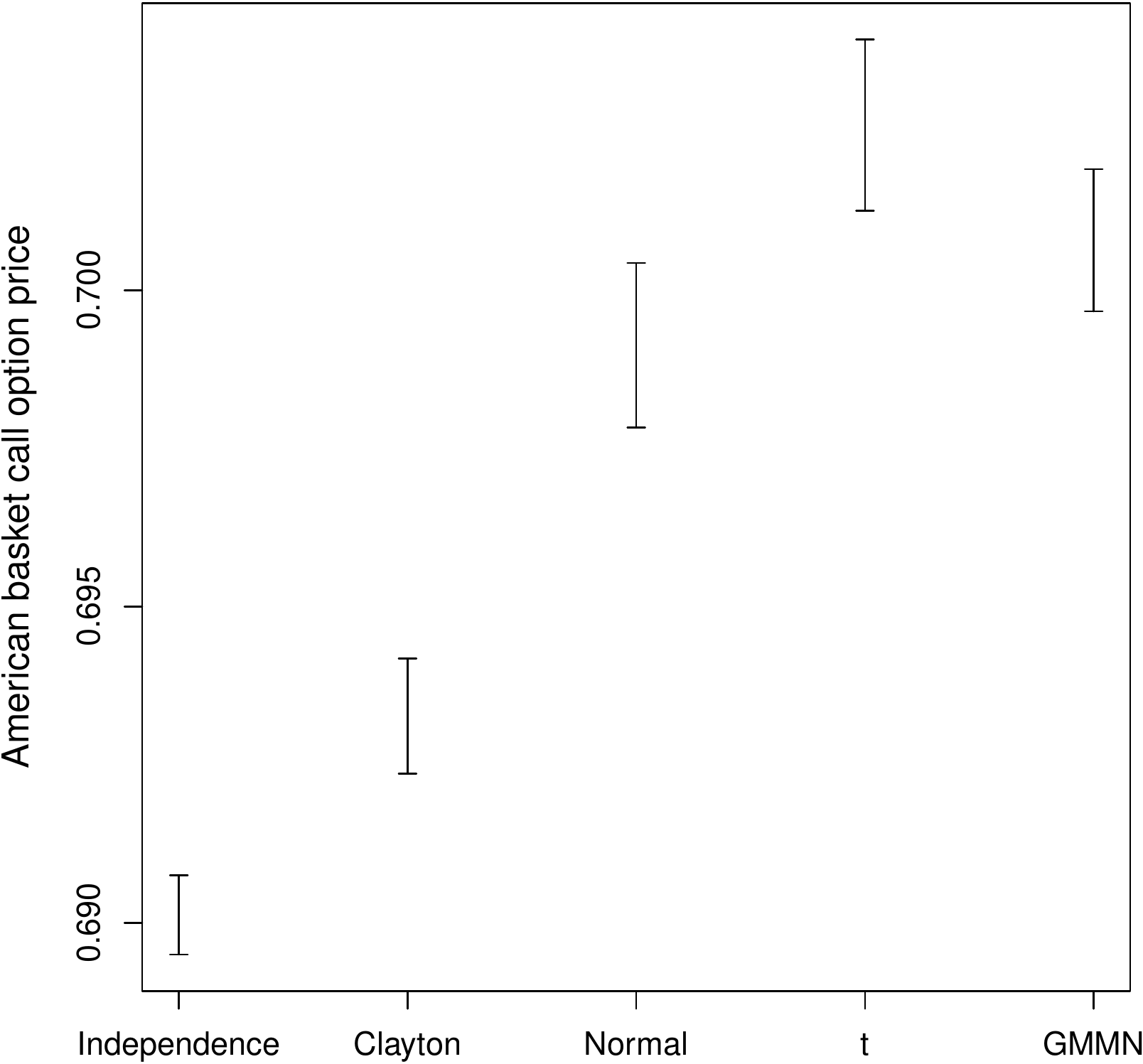}\hfill
\includegraphics[width=0.33\textwidth]{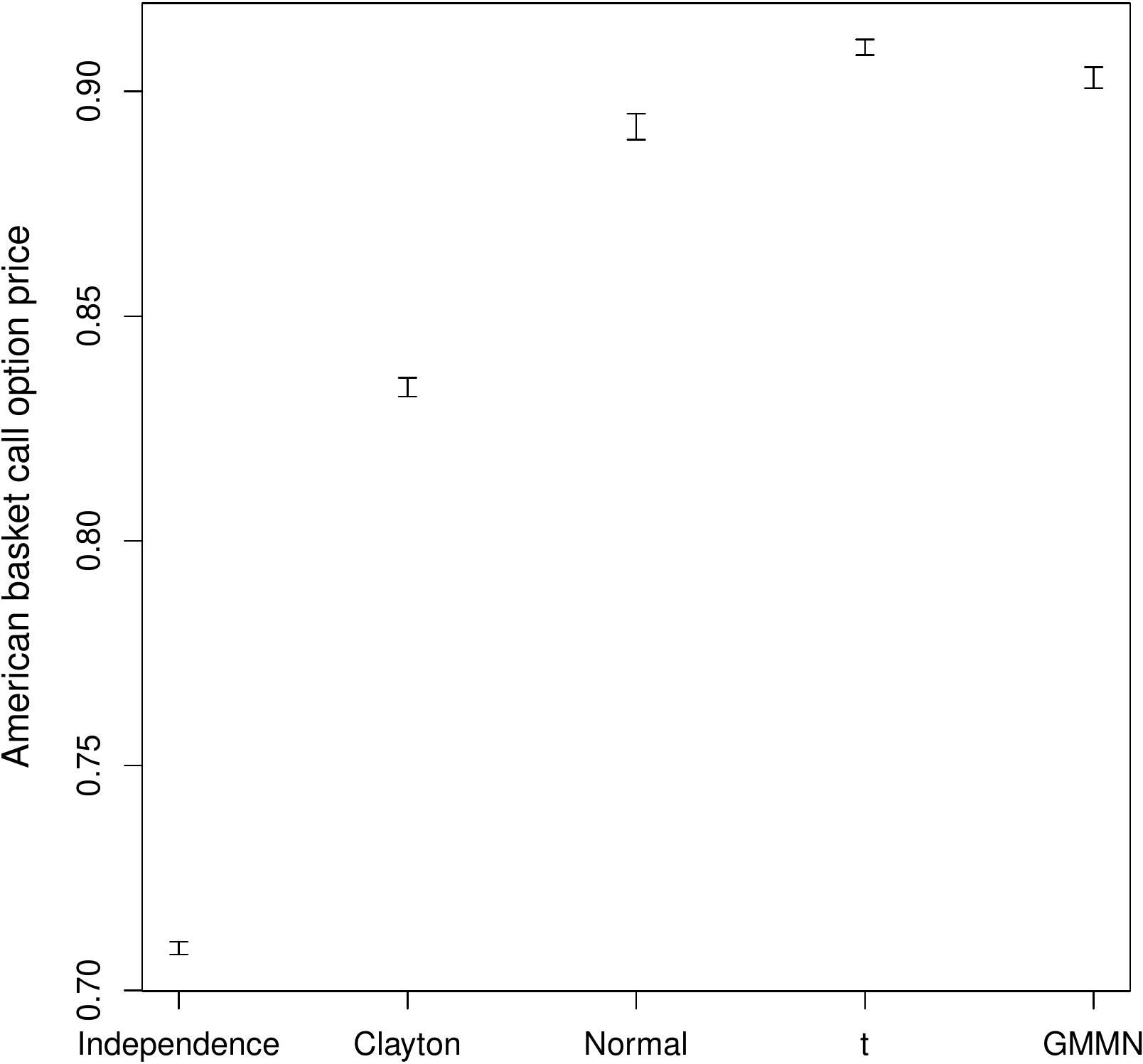}\hfill
\includegraphics[width=0.33\textwidth]{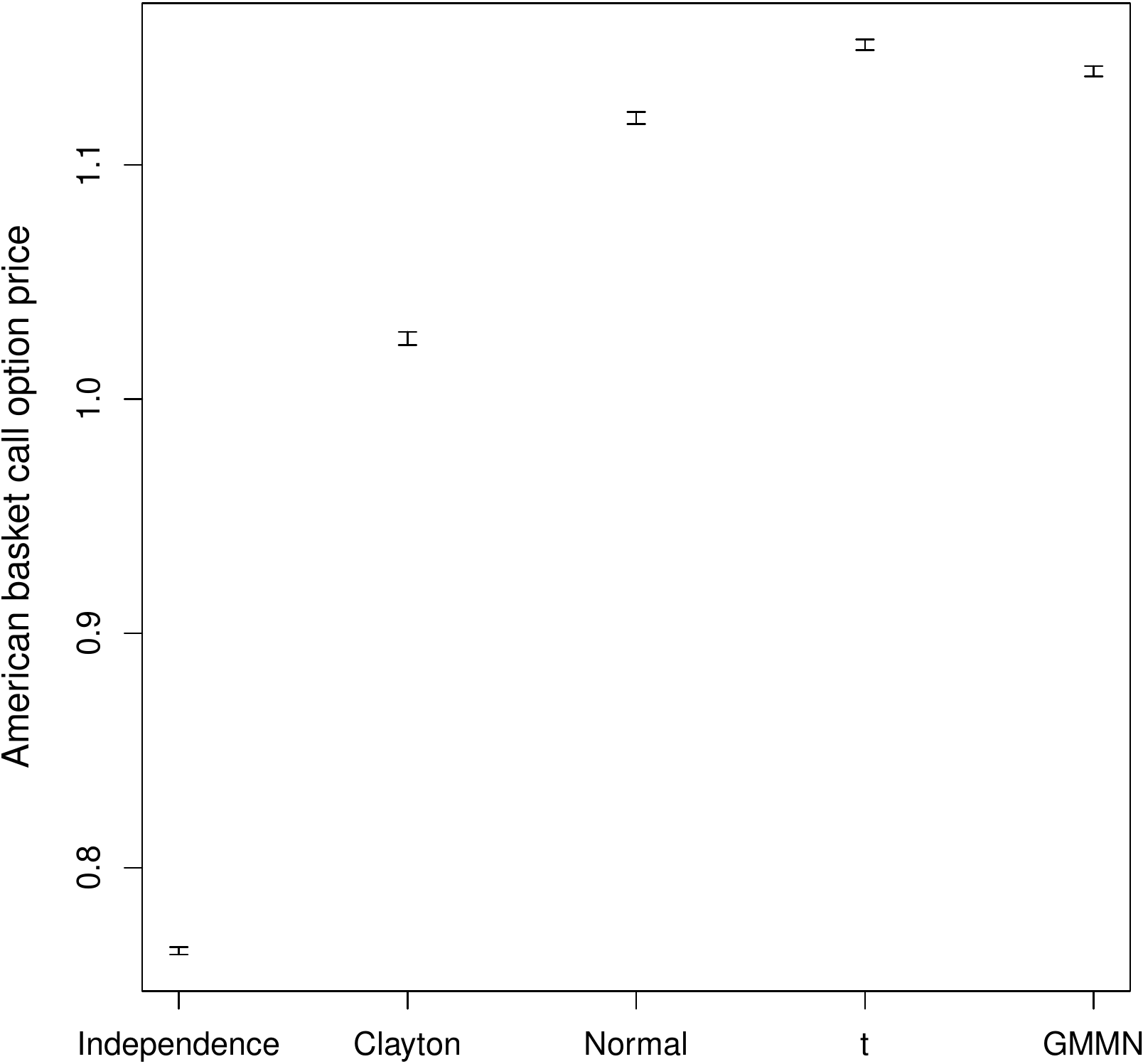}
\caption{95\%-confidence intervals of American basket call option prices with strike prices $K=79$ (top), $K=69$ (middle) and $K=81$ (bottom) based on $\nrep=25$ replications and $\npth=10\,000$ sample paths in each replication at $T=10$ (left), $T=50$ (middle) and $T=100$ (right) days to maturity based on portfolios of sizes $d=3$ (top), $d=5$ (middle) and $d=10$ (bottom). From these plots, we see that the option prices produced by GMMNs are similar to those produced by the $t$ and normal copulas, but still not the same as either of them.}\label{fig:option:CIs}
\end{figure}

Figure~\ref{fig:option:CIs} shows (Wald-type) 95\%-confidence intervals of the
American basket call option price for all dependence models for different
maturities (columns) and portfolios (rows) based on $\nrep=25$ replications. We
see that the GMMN leads to option prices that are similar to those given by the
normal and the $t$ copulas. This is not surprising as those two parametric
copulas are widely used and generally believed to be ``not too wrong'' for these
types of financial data, unlike the independence or the Clayton copulas which
are ``clearly wrong''. However, we also notice that the pricing provided by
the GMMN is often a few cents different from the ones given by the normal and $t$
copulas. While we do not know what the true dependence model and hence what the
true prices are, given that the GMMN fits better
(Figure~\ref{fig:gof:option}) our results here suggest that, more likely than
not, the normal and $t$ copulas do not correctly capture the underlying dependence
among these asset prices, either, and that those who are willing to so speculate
can exploit the potential arbitrage opportunities created by the GMMN.

\begin{figure}[htbp]
\centering
\includegraphics[width=0.33\textwidth]{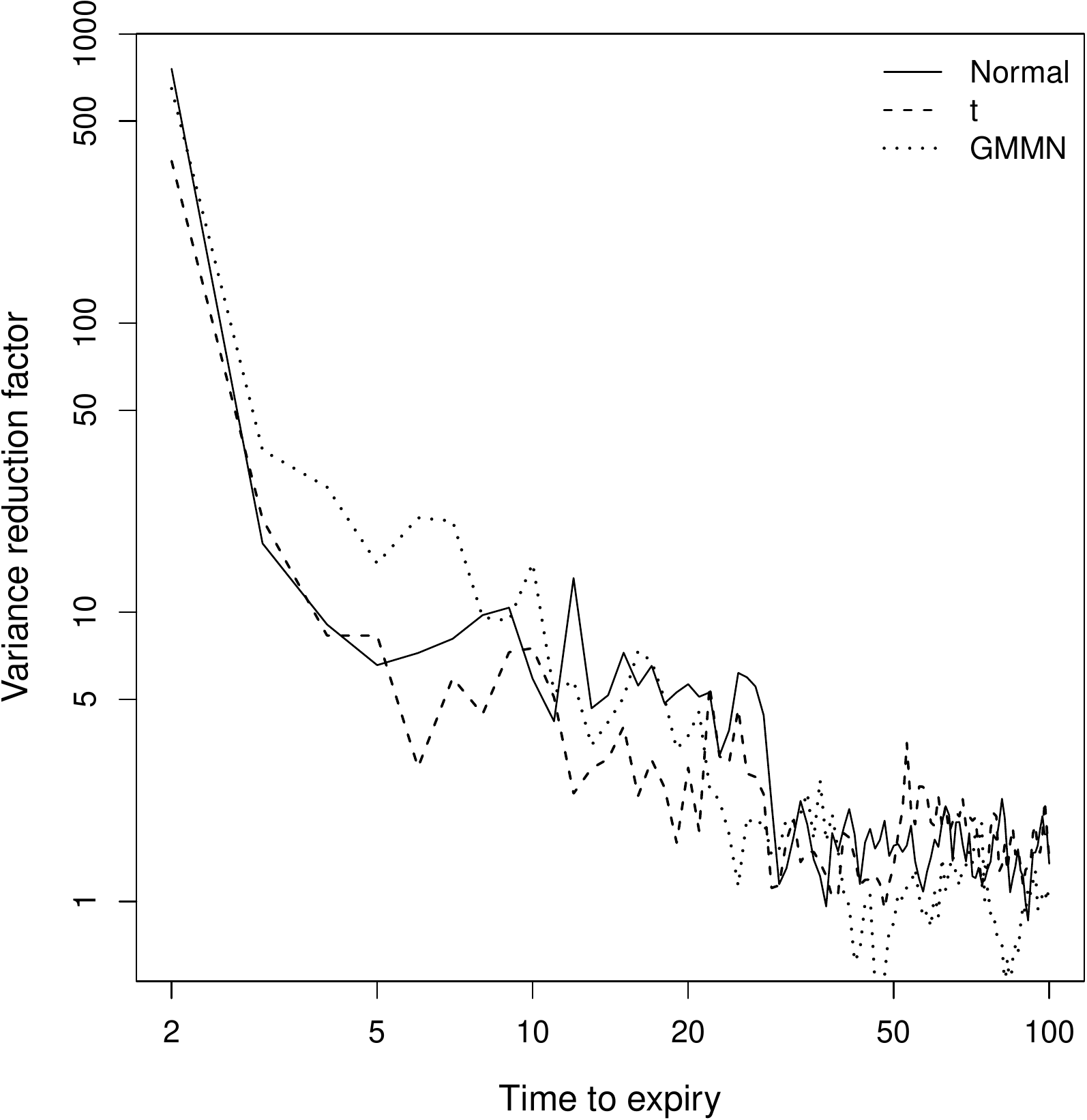}\hfill
\includegraphics[width=0.33\textwidth]{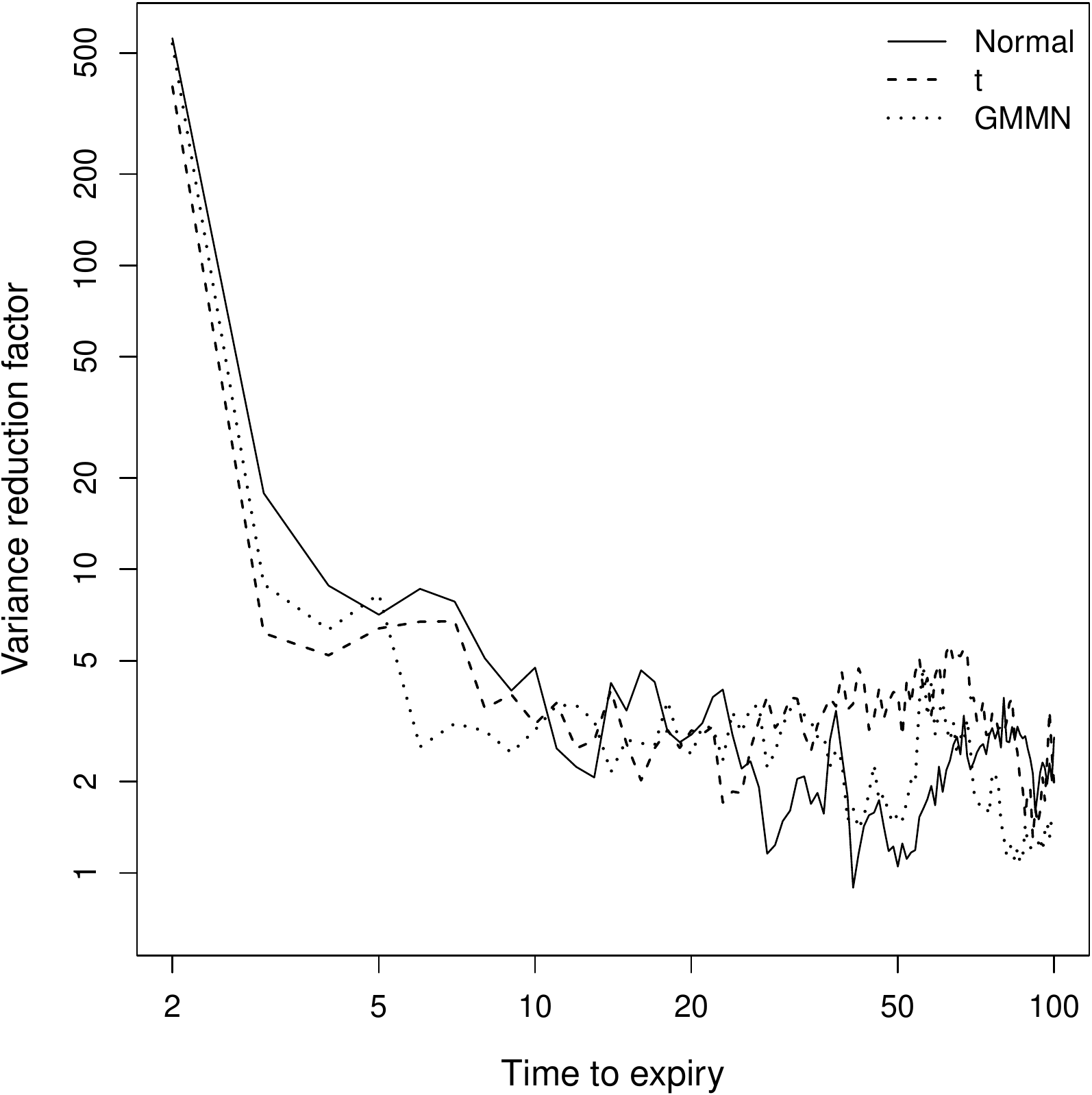}\hfill
\includegraphics[width=0.33\textwidth]{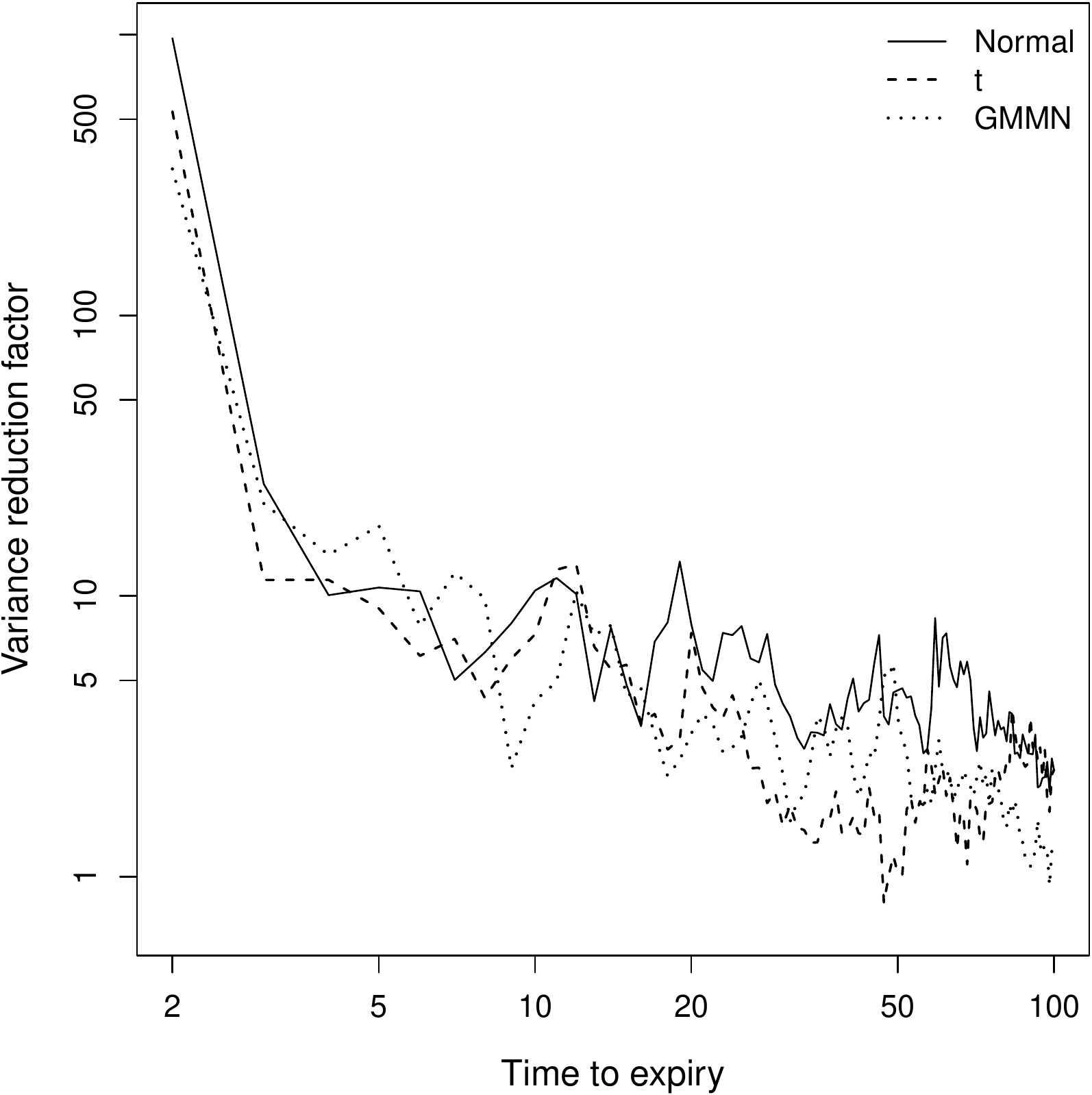}
\caption{Mean variance reduction factor estimates (computed over $\nrep=25$ replications and $\npth=10\,000$ paths in each replication) as a function of days to maturity for portfolios of sizes $d=3$ (left), $d=5$ (middle) and $d=10$ (right). From these plots, we see that the variance reduction effects are roughly similar for both the GMMNs and the two copula models and that this effect deteriorates when pricing options with longer times to maturity.}\label{fig:option:VRFs}
\end{figure}

In Figure~\ref{fig:option:VRFs} we focus on the best three models according to
Figure~\ref{fig:gof:option} and investigate the mean variance reduction factors
(determined based on $\nrep=25$ replications) when moving from pseudo-random to
quasi-random numbers. We see that for shorter times to maturity we get larger
variance reduction factors, but the effect deteriorates for longer times to
maturity. This is true across all considered dependence
models.

\section{Probabilistic forecasting for multivariate time series}
\label{sec:mts}

In this section, we consider the application of analyzing multivariate time series data $\bm{X}_t=(X_{t,1},\dots,X_{t,d})$ using the copula--GARCH approach \parencite{jondeaurockinger2006,patton2006}.

\subsection{Model}\label{sec:mts:model}
For simplicity, we model each individual time series $X_{t,j}$ as an $\ARMA(1,1)$--$\GARCH(1,1)$ process
\begin{align*}
X_{t_k,j}       &=\mu_{t_k,j} + \sigma_{t_k,j}Z_{k,j},\\
\mu_{t_k,j}     &=\mu_j+\phi_{j}(X_{t_{k-1},j}-\mu_{j}) +\gamma_{j}(X_{t_{k-1},j}-\mu_{t_{k-1},j}),\\
\sigma^2_{t_k,j}&=\omega_{j}+\alpha_{j}(X_{t_{k-1},j}-\mu_{t_{k-1},j})^2 +\beta_{j}\sigma^2_{t_{k-1},j},
\end{align*}
where $\omega_{j}>0$, $\alpha_{j},\beta_{j}\geq 0$, $\alpha_j+\beta_j<1$,
$|\phi_j|,|\gamma_j|<1$, and $\phi_j+\gamma_i\neq 0$ to guarantee a causal,
invertible and covariance stationary solution; see
\cite[Chapter~4]{mcneilfreyembrechts2015}. Of course, higher-order
$\ARMA$--$\GARCH$ processes can also be used, but that does not affect what we
are trying to demonstrate in this section.

For fixed $j$, the $Z_{k,j}$'s are iid
according to $F_{Z_j}$ across all time points $t_k$ with mean zero and unit
variance. In financial time series applications, it is common to model $F_{Z_j}$
as a standard normal, standardized $t$ or standardized skewed $t$
distribution. In the case of the latter, it is important to note that each
$F_{Z_j}$ is allowed to have a different degree of freedom.

These marginal $\ARMA$--$\GARCH$ models capture the serial dependence within
each individual time series. To capture the cross-sectional dependence between
individual series, copulas are used to model the distribution of
$(F_{Z_{1}}(Z_{k,1}),\dots,F_{Z_{d}}(Z_{k,d}))$.

\subsection{Estimation}
\label{sec:mts-estimation}

In the context of this application, \eqref{eq:remove} amounts to the process of estimating everything --- i.e., $\mu_j$, $\phi_{j}$, $\gamma_{j}$, $\omega_j$, $\alpha_{j}$, $\beta_{j}$ and (parameters of) $F_{Z_j}$ --- and then removing them from all the given data $X_{t_k,j}$, $k=1,\dots,\ntrn$, by
\begin{align*}
  \hat{Z}_{k,j}  &= (X_{t_k,j} - \hat{\mu}_{t_k,j})/\hat{\sigma}_{t_k,j}, \\
  \hat{U}_{k,j}&= \hat{F}_{Z_j}(\hat{Z}_{k,j}).
\end{align*}
These steps are known in the literature as deGARCHing. As in Section~\ref{sec:opt:price}, using the resulting $\hat{U}_{k,j}$, we can either estimate a parametric copula model, $\hat{C}_{\text{PM}}$, or train a nonparametric GMMN, $\hat{C}_{\text{NN}}$. In the case of the former, we call this the \emph{copula--GARCH approach}; in the case of the latter, we call it the \emph{GMMN--GARCH approach}.

\subsection{Forecast}
\label{sec:mts:forecast}

To produce probabilistic forecasts, we must now simulate each $X_{t,j}$ process forward, according to \eqref{eq:simulate}. Suppose we have observed the process up to and including time $t_k$, and would like to forecast $h$ periods ahead; that is, we'd like to independently simulate $\npth$ paths forward and generate $\hat{X}^{(i)}_{t_{k'},j}$ for $i=1,\dots,\npth$ and $k'=k+1,\dots,k+h$.

For fixed $i$ and $k'$, this is achieved for the GMMN--GARCH approach by first generating $\bm{U}^{(i)}_{k'} = (U^{(i)}_{k',1},\dots,U^{(i)}_{k',d})$ with either Algorithm~\ref{algorithm:GMMN:prng2} (for pseudo-random samples) or Algorithm~\ref{algorithm:GMMN:qrng2} (for quasi-random samples), then letting
\begin{align*}
  Z^{(i)}_{k',j} = \hat{F}^{-1}_{Z_j}(U^{(i)}_{k',j}),
\end{align*}
and finally simulating $\hat{X}^{(i)}_{t_{k'},j}$ according to
\begin{align*}
\hat{\mu}^{(i)}_{t_{k'},j}
&=\hat{\mu}_j+\hat{\phi}_{j}(\hat{X}^{(i)}_{t_{{k'}-1},j}-\hat{\mu}_j)
  +\hat{\gamma}_{j}(\hat{X}^{(i)}_{t_{{k'}-1},j}-\hat{\mu}^{(i)}_{t_{{k'}-1},j}),\\
\hat{\sigma}^{2^{(i)}}_{t_{k'},j}
&=\hat{\omega}_j
+\hat{\alpha}_{j}(\hat{X}^{(i)}_{t_{k'-1},j}-\hat{\mu}^{(i)}_{t_{k'-1},j})^2
+\hat{\beta}_{j} \hat{\sigma}^{2^{(i)}}_{t_{k'-1},j}, \\
\hat{X}^{(i)}_{t_{k'},j}
&=\hat{\mu}^{(i)}_{t_{k'},j} + \hat{\sigma}^{2^{(i)}}_{t_{k'},j} Z^{(i)}_{k',j},
\end{align*}
where, for $k' \leq k$, we simply set $\hat{X}^{(i)}_{t_{k'},j}=X_{t_{k'},j}$,
$\hat{\sigma}^{2^{(i)}}_{t_{k'},j}=\hat{\sigma}^{2}_{t_{k'},j}$, and
$\hat{\mu}^{(i)}_{t_{k'},j}=\hat{\mu}_{t_{k'},j}$ for all $i$. Notice that it's possible to do this at $t_k > t_{\ntrn}$. Then, all observed quantities up to and including $\bm{X}_{t_k}$ are used to make forecasts, but we do not re-estimate anything that has already been estimated in Section~\ref{sec:mts-estimation} using only observed quantities up to and including $\bm{X}_{t_{\ntrn}}$.

The collection of simulated paths,
\begin{align*}
\{\hat{\bm{X}}_{t_{k+1}}^{(i)}, \hat{\bm{X}}_{t_{k+2}}^{(i)}, \dots,
  \hat{\bm{X}}_{t_{k+h}}^{(i)} \,|\, \mathcal{F}^{(i)}_{t_k}\}_{i=1}^{\npth},
\end{align*}
encode an \emph{empirical predictive distribution} (EPD) at each time point $t_{k+1},\dots,t_{k+h}$, from which various probabilistic forecasts can be made --- for example, we can forecast $\P(\bm{X}_{t_{k+h}}\in A)$ by $(1/\npth)\sum_{i=1}^{\npth} \ID(\hat{\bm{X}}_{t_{k+h}}^{(i)} \in A)$ for any given $A\subset\IR^d$. However, below when we assess $h$-period-ahead EPDs (made at time $t_k$), we will only be comparing $\{\hat{\bm{X}}_{t_{k+h}}^{(i)}|\mathcal{F}^{(i)}_{t_k}\}_{i=1}^{\npth}$ with $\bm{X}_{t_{k+h}}$, not any of the ``intermediate'' forecasts made along the way at $t_{k+1},\dots,t_{k+h-1}$.

\subsection{Application}
\label{sec:mts-data}

We illustrate with two exchange rate data sets: a US dollar (USD) data set consisting of daily exchange rates of Canadian dollar (CAD), Pound sterling (GBP), Euro (EUR), Swiss Franc (CHF) and Japanese yen (JPY) with respect to the USD; and a GBP data set consisting of daily exchange rates of CAD, USD, EUR, CHF, JPY and the Chinese Yuan (CNY) with respect to the GBP. For further details regarding both the data sets, see the \R\ package \texttt{qrmdata}.

In particular, we consider these multivariate time series from $t_1$=2000-01-01 to $t_{\nall}$=2015-12-31, treating data up to $t_{\ntrn}$=2014-12-31 as the training set and the remainder as a held-out test set.

The distribution $F_{Z_j}(z_i) = t_{\nu_j}(z_j\sqrt{\nu_j/(\nu_j-2)})$ is chosen to be the scaled $t$-distribution for all $j=1,\dots,d$. After the steps in Section~\ref{sec:mts-estimation}, we fit three dependence models to $\hat{U}_{k,j}$, $k=1,\dots,\ntrn$, $j=1,\dots,d$: as $\hat{C}_{\text{PM}}$, a normal copula and a $t$-copula, both with unstructured correlation matrices; and as $\hat{C}_{\text{NN}}$, a GMMN with the same architecture and hyperparameters as described in \cite{hofertprasadzhu2021}. In addition, we use batch normalization and dropout regularization (with a dropout rate of $0.3$) to help control for overfitting while training.

\begin{figure}[htbp]
	\centering
	\includegraphics[width=0.49\textwidth]{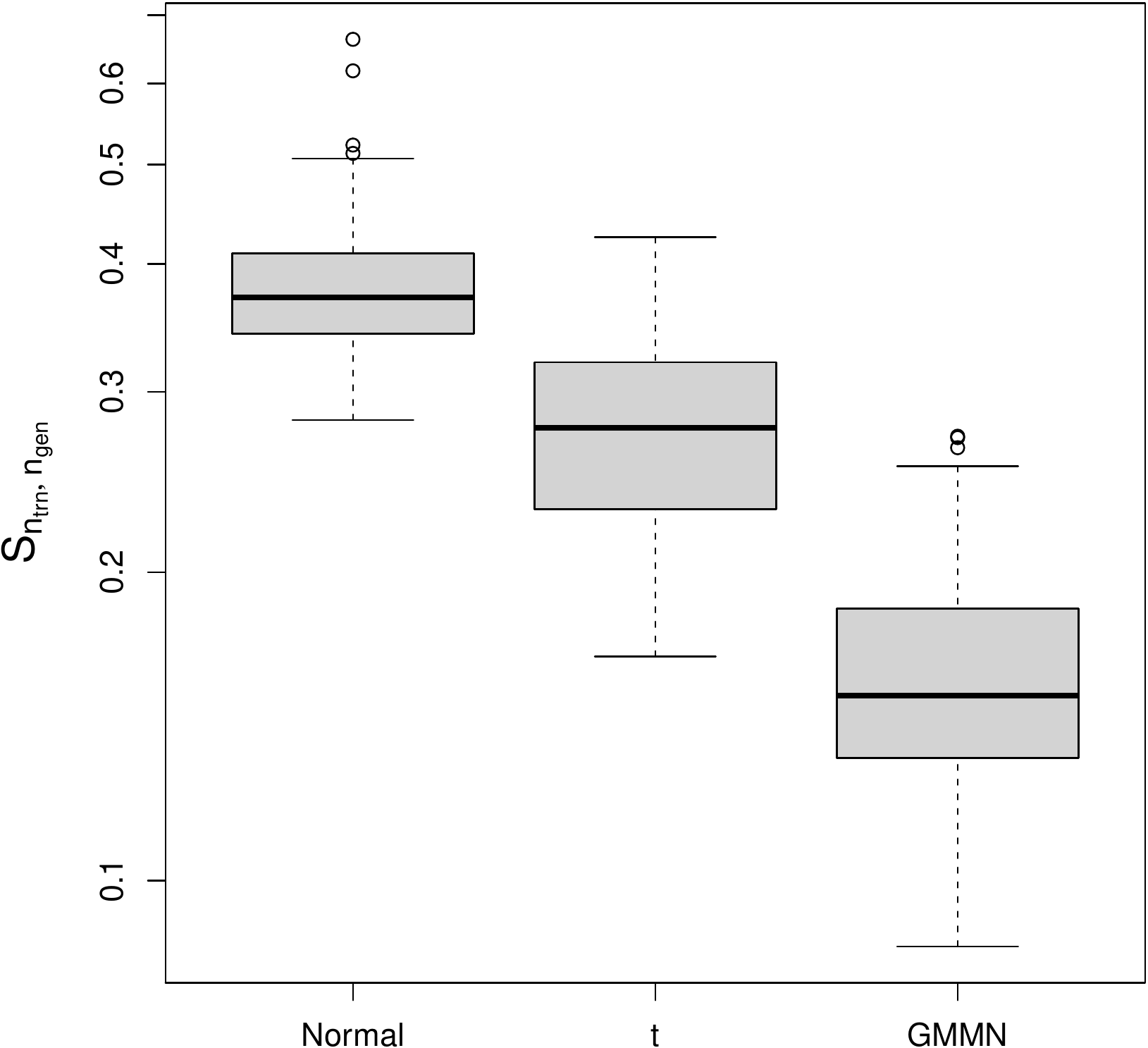}\hfill
	\includegraphics[width=0.49\textwidth]{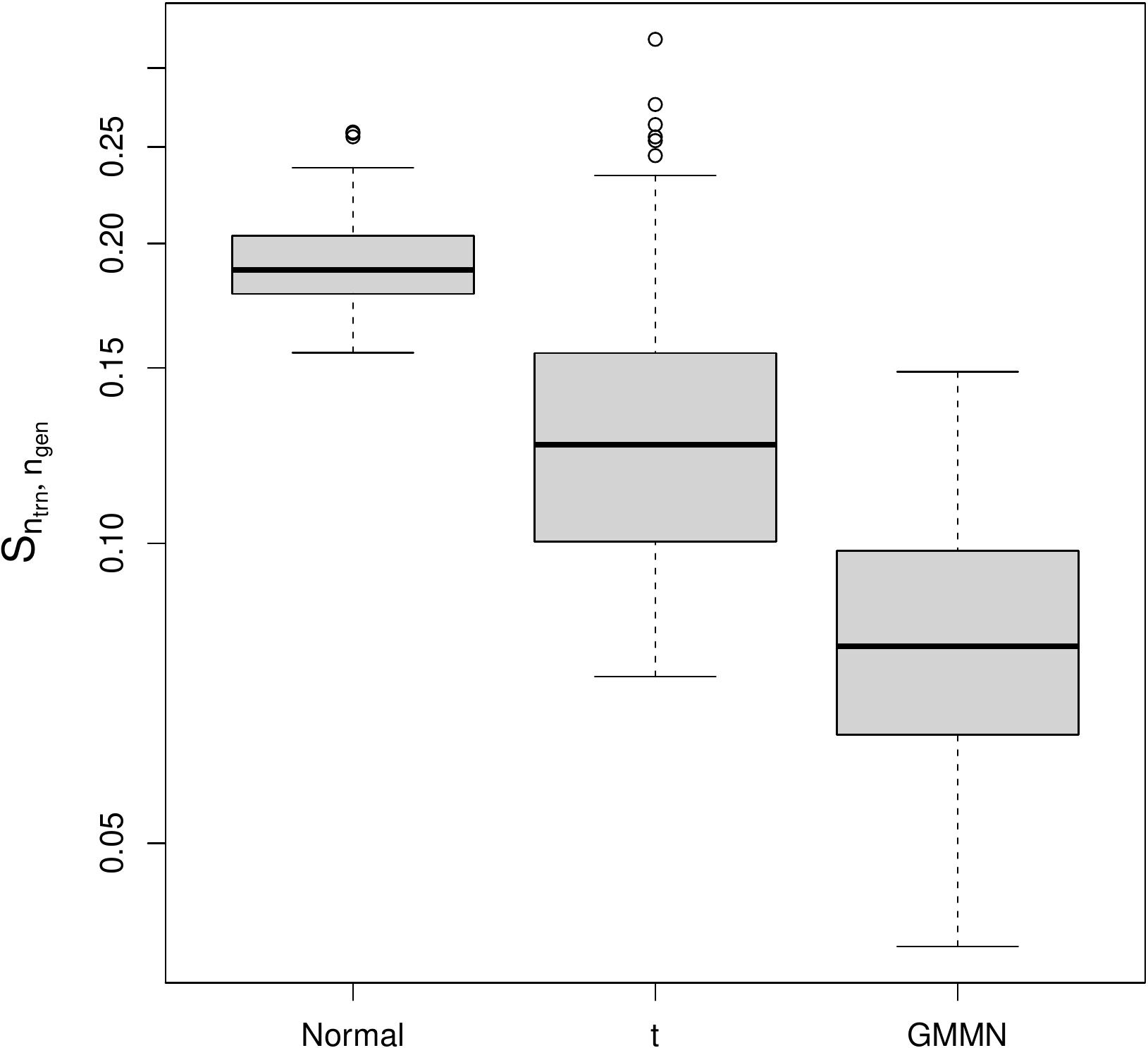}
	\caption{Box plots based on $\nrep=100$ realizations of
            $S_{\ntrn,\ngen}$ computed for the FX USD (left) and FX GBP (right)
            data sets of size $\ntrn=5478$ with dimensions $d=5$ and $d=6$,
            respectively, and for each fitted dependence model using a
            pseudo-random sample of size $\ngen=10\,000$. From these box plots,
            we can see that the GMMNs provide a much better fit than the two
            parametric copulas for both data sets.}\label{fig:gof:mts}
\end{figure}

To assess the fit of these dependence models, we use a Cram\'{e}r-von-Mises type
test statistic presented in~\eqref{eq:cvm}. From Figure~\ref{fig:gof:mts} which displays box plots of $S_{\ntrn,\ngen}$ for the
different models based on $\nrep=100$ repetitions,
we can see that GMMNs clearly provide the best fit across both FX USD and FX GBP data sets.

The key question, though, is whether better fits translate to better predictions.
For fixed horizon $h$, we can produce an $h$-day-ahead EPD $\{\hat{\bm{X}}^{(i)}_{t_{k+h}}|\mathcal{F}^{(i)}_{t_k}\}_{i=1}^{\npth}$ at every $t_k=t_{\ntrn},\dots,t_{\nall-h}$ in the test period. To assess the quality of the EPDs produced by the copula--GARCH and the GMMN--GARCH approaches, we compare the respective EPD with the actual realization $\bm{X}_{t_{k+h}}$ in the held-out test set using the \emph{variogram score} introduced by \cite{scheuerer2015}, which, in our context, assesses if the EPD is biased for the distance between any two component samples. For a single numeric summary, we work with an average variogram score (of order $r$) over the entire period $t_{\ntrn},\dots,t_{\nall-h}$
\begin{align}
  &\phantom{{}={}}\text{AVS}^{r}_{h}\notag\\
  &=\frac{1}{\nall-h-\ntrn} \sum_{k=\ntrn}^{\nall-h}
\sum_{j_1=1}^{d}\sum_{j_2=1}^{d} \biggl(|X_{t_{k+h},j_1}-X_{t_{k+h},j_2}|^r- \frac{1}{\npth}\sum_{i=1}^{\npth}
 |\hat{X}^{(i)}_{t_{k+h},j_1}-\hat{X}^{(i)}_{t_{k+h},j_2}|^r\biggr)^2. \label{eq:average:VS}
\end{align}
\cite{scheuerer2015} numerically demonstrated that, by focusing on pairwise distances between component samples, this metric discriminates well between various dependence structures. They also stated that a typical choice of the variogram order might be $r=0.5$, but noted in their concluding remarks that smaller values of $r$ could potentially yield more discriminative metrics when dealing with non-Gaussian data, which is why we choose to work with $r=0.25$.

\begin{figure}[htbp]
  \centering
  \includegraphics[width=0.33\textwidth]{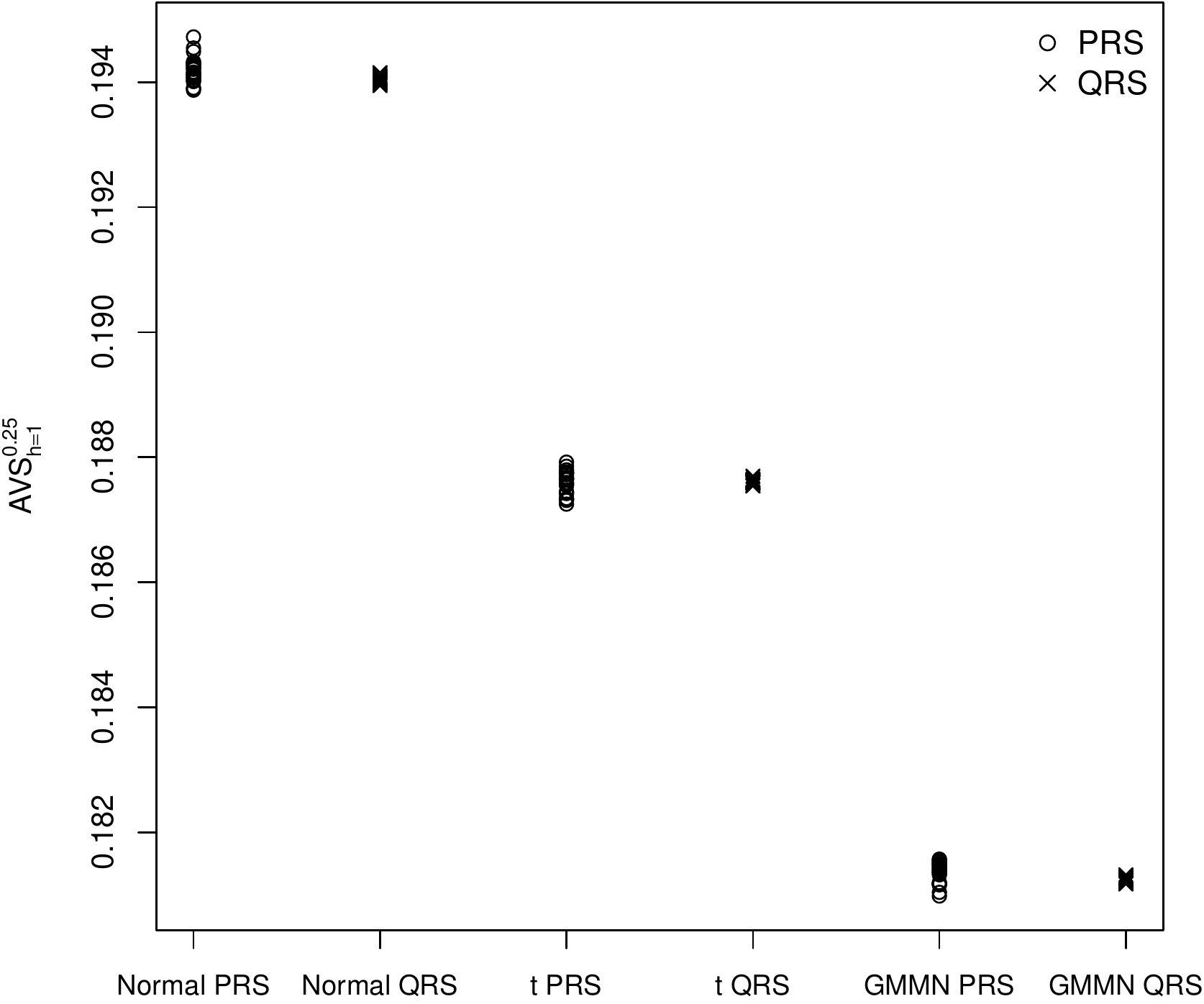}\hfill
  \includegraphics[width=0.33\textwidth]{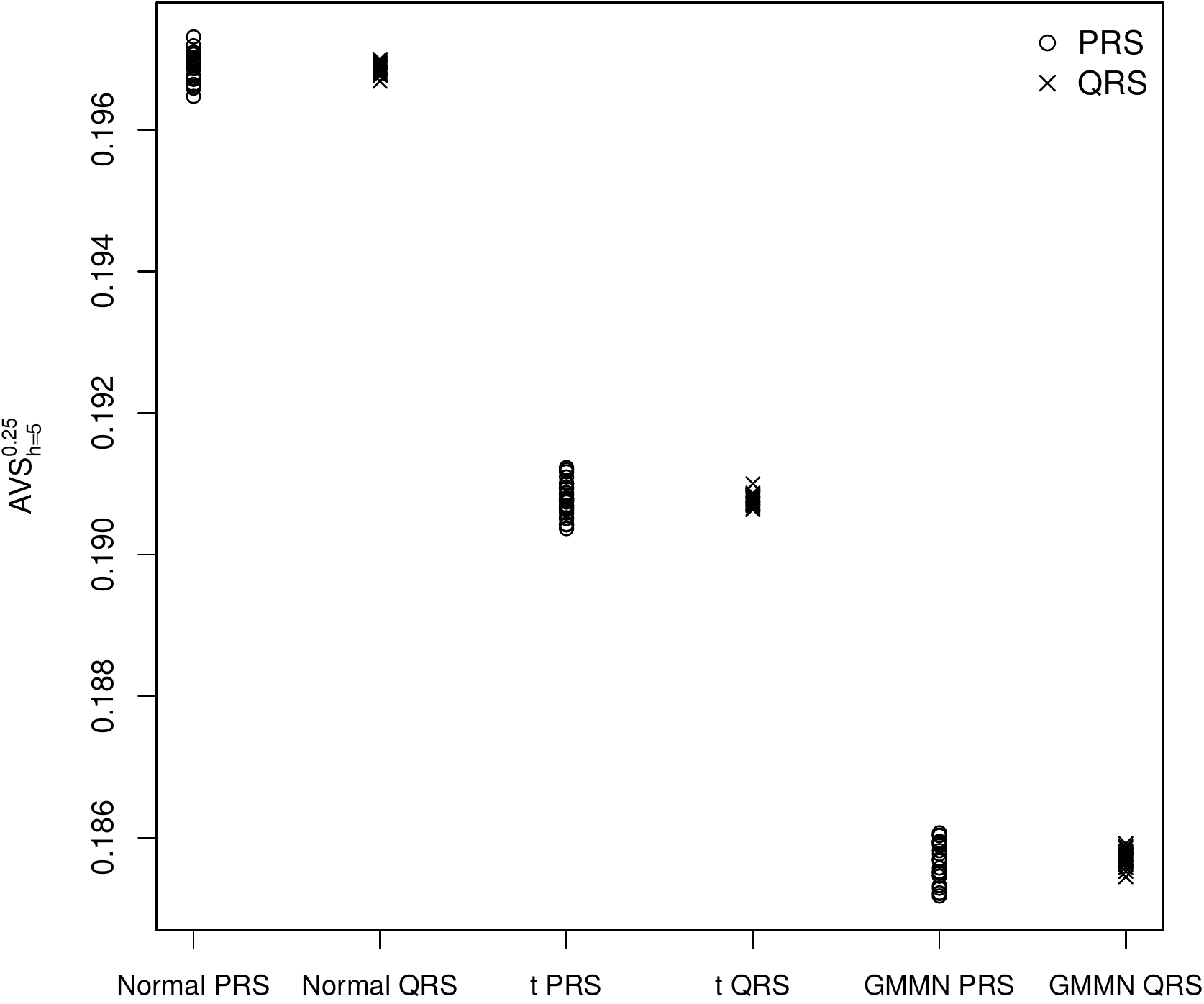}\hfill
  \includegraphics[width=0.33\textwidth]{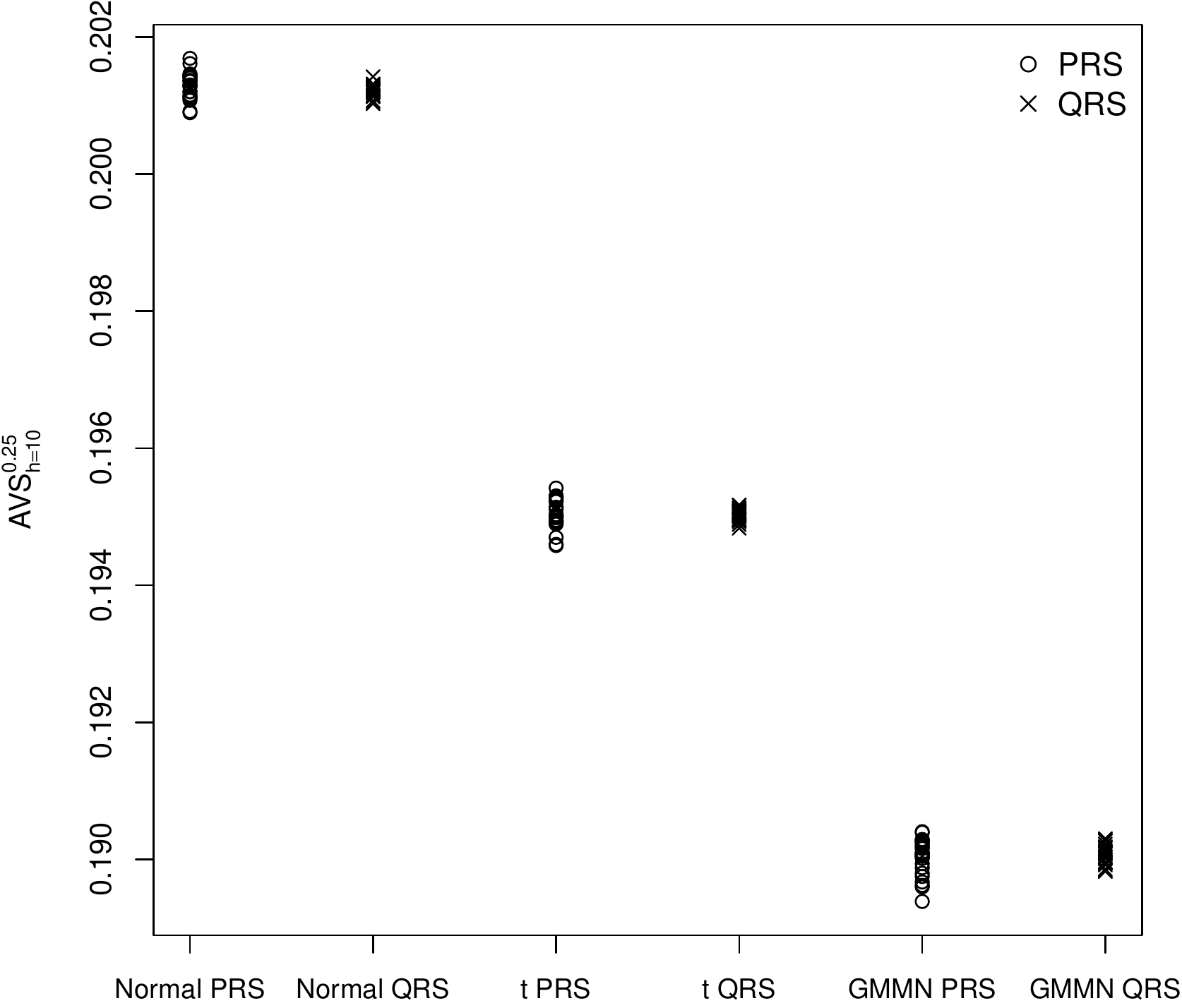}\\[2mm]
  \includegraphics[width=0.33\textwidth]{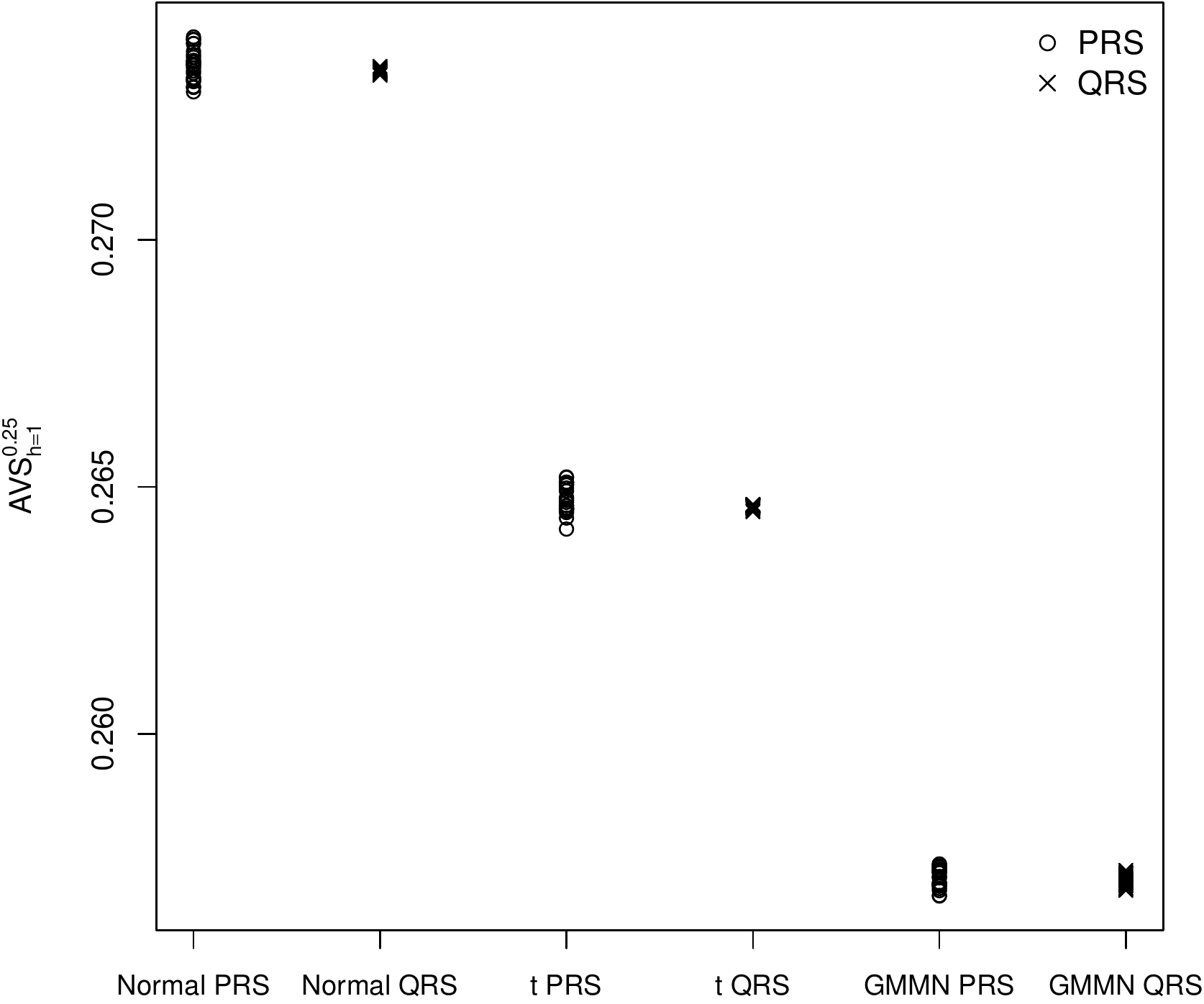}\hfill
  \includegraphics[width=0.33\textwidth]{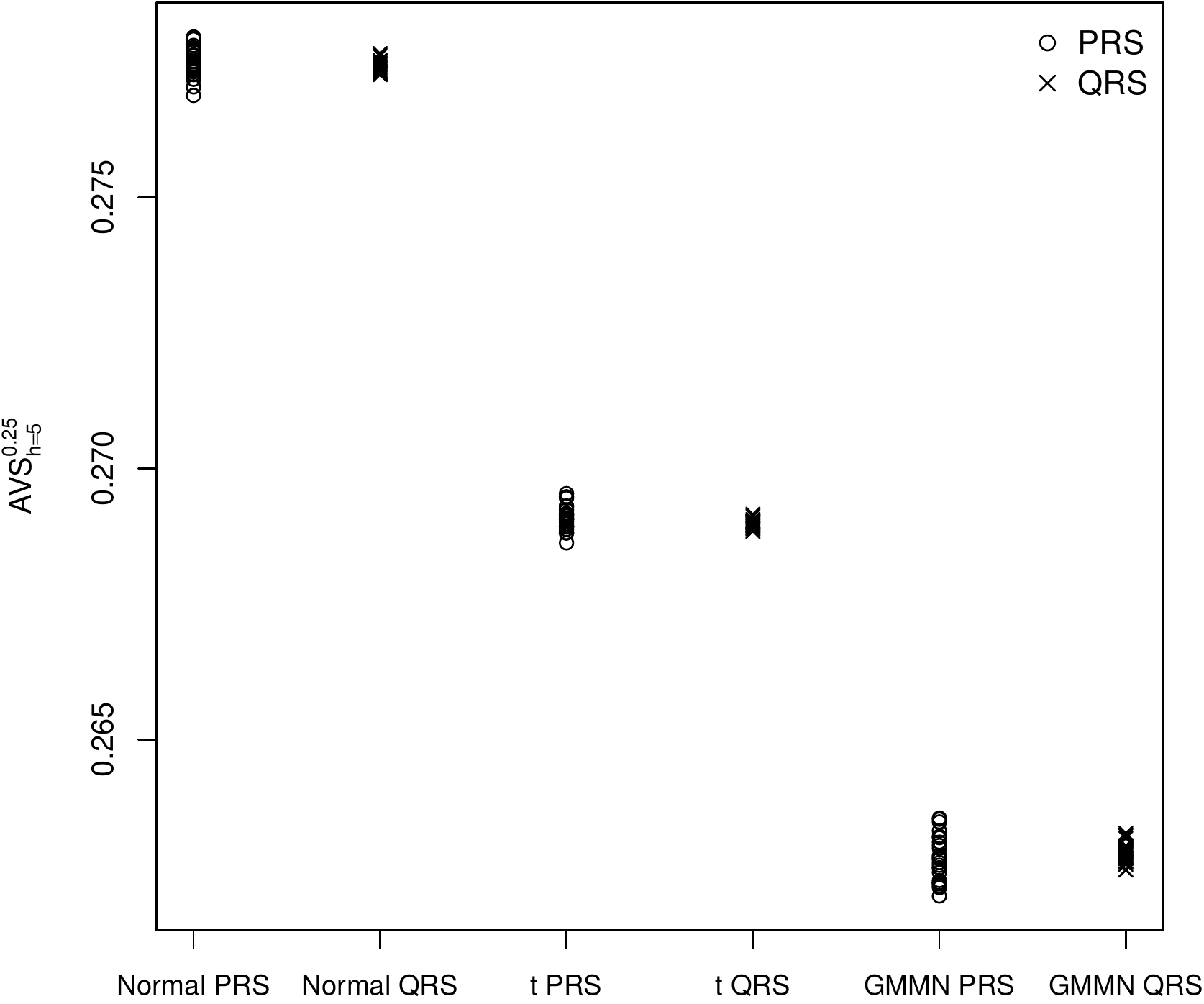}\hfill
  \includegraphics[width=0.33\textwidth]{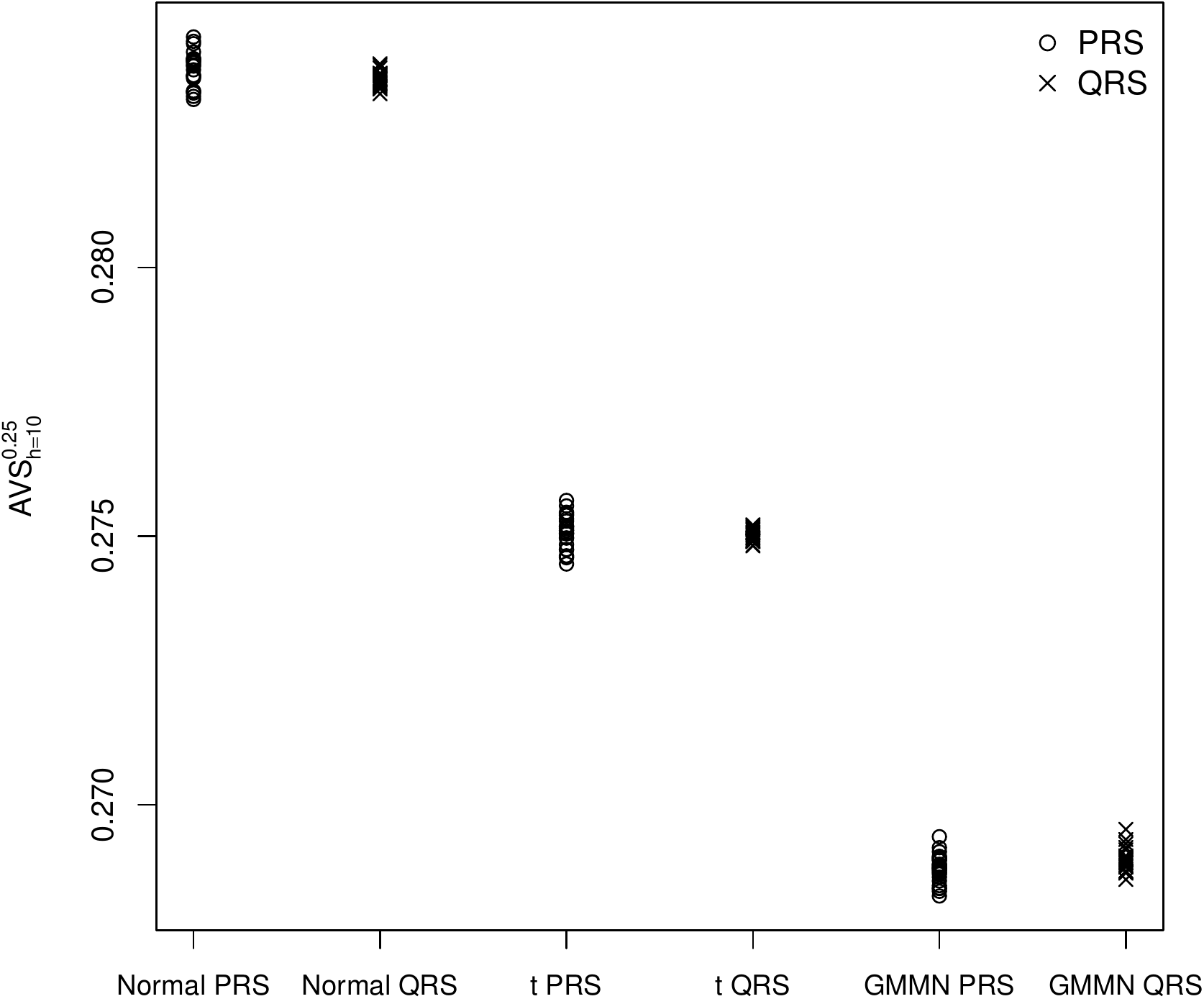}\hfill
  \caption{Replications of $\nrep=25$ average variogram scores
      $\text{AVS}^{0.25}_{h}$ based on $\npth=1000$ simulated paths in each
      replication, for $h=1$ (left), $h=5$ (middle) and $h=10$ (right) using
      pseudo-random as well as quasi-random samples from normal copulas, $t$
      copulas, and GMMNs for the FX USD (top) and FX GBP (bottom) data
      sets. From these plots, we observe that GMMN--GARCH models yield smaller
      variogram scores and hence better EPDs when compared to various
      copula--GARCH models. Furthermore, we observe a clear variance reduction effect when
      using quasi-random samples to compute the variogram scores across multiple
      replications.}\label{fig:mts-AVS}
\end{figure}

Figure~\ref{fig:mts-AVS} shows that, for both the USD and the GBP data sets, the GMMN--GARCH approach has produced better EPDs (smaller variogram scores) overall for held-out realizations in the test set. Moreover, the resulting variogram scores are also more stable (less variation) over replications of the same experiment when quasi-random (as opposed to pseudo-random) samples are used.

\section{Conclusion}
We suggested GMMNs as cross-sectional dependence models for multivariate discrete-time
stochastic processes. As examples, we considered discretized geometric
Brownian motions with an application to pricing American basket call options
under dependence, as well as $\ARMA$--$\GARCH$ models with an application to obtain
predictive distributions.  These examples have demonstrated two advantages of
GMMNs as dependence models. First, they provide more flexible dependence models than parametric copulas, which make a difference when estimating quantities of interest such as option prices and making probabilistic forecasts. Second, they come with a ``built-in'' option to generate quasi-random samples and thus allow us to obtain a variance reduction effect without additional effort.

\printbibliography[heading=bibintoc]
\end{document}

%
%
%
%
